\documentclass{siamltex}
\usepackage{epstopdf}

\usepackage{latexsym, graphicx, epsfig}
\usepackage{url}
\usepackage{color}
\usepackage{array}
\usepackage{amsfonts,amsmath,amssymb}
\usepackage{verbatim}
\usepackage{subfigure}
\usepackage[ruled, linesnumbered]{algorithm2e}
\usepackage{enumerate}
\usepackage{float}

\newtheorem{remark}{Remark}[section]

\newcommand{\lv}{\lVert}
\newcommand{\rv}{\rVert}

\title{Bifidelity data-assisted  neural networks in nonintrusive reduced-order modeling}
\author{    Chuan Lu\thanks{
        Department of Mathematics, University of Iowa, Iowa City, IA 52242 Email: chuan-lu@uiowa.edu}     \and
Xueyu Zhu\thanks{Corresponding author, Department of Mathematics, University of Iowa, Iowa City, IA 52242 Email: xueyu-zhu@uiowa.edu.}     
                 }

\begin{document}

\maketitle

\begin{abstract}
{
In this paper, we present a new nonintrusive reduced basis  method 
when a cheap low-fidelity model and  expensive high-fidelity model are
 available. The method relies on proper orthogonal decomposition (POD) to generate the high-fidelity reduced basis and a shallow multilayer perceptron to learn the high-fidelity reduced coefficients. In contrast to other methods, one distinct feature of the proposed method is to  incorporate the features extracted from the low-fidelity data as the input feature, this approach not only improves the predictive capability of the neural network but also enables the decoupling the high-fidelity simulation from the online stage. 
Due to its nonintrusive nature, it is applicable to general parameterized problems. We also provide several numerical examples to illustrate the effectiveness and performance of the proposed method.} 
\end{abstract}



\pagestyle{myheadings}
\thispagestyle{plain}


\section{Introduction}
\label{sec:intro}


 Parameterized partial differential equations (PDEs) arise in many complex scientific and engineering applications. A common task in such applications requires solving the underlying PDE efficiently and accurately for a large number of parameter points in the parameter space, which poses a huge computational challenge, particularly for  large-scale problems. To address this challenge, reduced-order modeling or model reduction techniques \cite{quarteroni2015reduced,hesthaven2016certified} have been  proven to be successful for many practical problems with low intrinsic dimension, including electromagnetic scattering \cite{chen2012certified}, multiscale simulations \cite{hesthaven2015reduced},  and uncertainty quantification \cite{chen2016model}, to name a few.  

As one of major  model reduction techniques, classical projection-based model reduction algorithms, such as proper orthogonal decomposition (POD) based method and reduced  basis method (RBM) \cite{rozza2007reduced}, generally follows an offline-online paradigm \cite{maday2006reduced}. At the offline stage, a set of reduced basis is built from a collection of full-order simulation results. 
During the online stage, for a new parameter value, the reduced model is constructed as a linear combination of the pre-computed reduced basis, where the expansion coefficients are computed by projecting  the full-order equation onto the reduced approximation space \cite{buffa2012priori}. 
Despite its success for many applications, this coupling between the full-order model and the online stage  requires major rewrites of the  sophisticated original legacy solver of the full order model. Additionally, it  causes  computational inefficiency for nonlinear problems, where the online computational complexity of the nonlinear
term remains high due to its dependence on the degrees of freedom (DOFs) of the full-order solution, instead of the dimension of reduced approximation space. 

To tackle these challenges, non-intrusive methods such as \cite{xiao2015non,xiao2016non} have been developed to construct a surrogate of the high-fidelity reduced coefficients so that  the reduced coefficients can be recovered without requiring a projection of the full-order model. More specifically, the full-order simulations are only required for basis generation during the offline stage.  During the online stage, the reduced coefficients are  recovered by interpolation over the parameter space. 
Even though this approach enables decoupling between the full-order model and the online stage, the interpolation approaches are prone to fail in complex applications, where the reduced coefficient has a highly nonlinear dependency on the parameters  \cite{amsallem2010interpolation}.

In a recent work \cite{hesthaven2018non}, a novel non-intrusive reduced basis method combining POD and neural network  was proposed, referred to as POD-NN. In this method, neural networks, especially a shallow multi-layer perceptron (MLP) is used to approximate the POD coefficients of high-fidelity data. Instead of using a nonadapted  interpolation basis set, neural networks provide a data-driven approach to approximate the mapping from the physical model parameter space to the high-fidelity reduced coefficient space. For nonlinear parameterized problems, it shows a  superior performance 
compared to traditional POD-Galerkin based methods. Nevertheless, it is important to note that the  input features of POD-NN are the physical model parameters, which are not data-dependent and might not be strongly informative.  While it would be desirable to utilize the input feature from the high-fidelity data set to train the network at the offline stage, this is still problematic since 
the online stage still requires a full-order  (high-fidelity)  simulation to generate the high-fidelity feature. 
Therefore, utilizing high-fidelity features is not practical for many scientific and engineering applications with expensive high-fidelity solvers. 

It is worth noting that the aforementioned methods are based on single fidelity solver of the full-order model, which is usually computationally expensive. In many practical problems, there often exists models with different fidelities \cite{alexandrov2000optimization,sun2011multi,cutler2014reinforcement}. 
Although the low-fidelity models are inaccurate, they can still mimic important behaviors of the underlying problem with a much lower computational cost. 
Many recent works in computational science and engineering community suggested that it is advantageous to combine low-fidelity and high-fidelity models together to improve computational accuracy or efficiency  under different contexts, such as uncertainty quantification \cite{narayan2014stochastic,zhu2014computational,zhu2017multi,peherstorfer2016multifidelity,hampton2018practical,perdikaris2015multi,yang2018bifidelity}, optimization \cite{forrester2007multi,robinson2008surrogate, leifsson2010multi}. 
Much of multifidelity research work in machine learning have been focused on  optimization setting \cite{cutajardeep,kandasamy2016gaussian,pmlr-v80-sen18a,hu2019multi} as well as the ensemble learning \cite{dietterich2000ensemble,zhou2012ensemble}. In general, the main differences between those multifidelity algorithms lie in the problem setting and the way to integrate the multifidelity model or data together.

Motivated by the recent developments in multifidelity modeling and reduced order modeling~\cite{hesthaven2018non}, we propose a new nonintrusive reduced basis method when  a cheap low-fidelity model and expensive high-fidelity model are available. In this work, we train a two-hidden-layer perceptron to approximate the high-fidelity reduced coefficients,  referred to as Bi-Fidelity data-assisted Neural Network (BiFi-NN).  
More specifically, besides using the original physical model parameter itself as the input feature  suggested in \cite{hesthaven2018non}, we also augment additional  features extracted from  the cheap low-fidelity data as the input feature. 
These new features are not only data-dependent, but also encode more prior information into neural networks.  Additionally, it decouples the high-fidelity simulations from the online stage. 
 This provides a competitive alternative to existing nonintrusive reduced basis methods that produce accurate reduced solutions with an  affordable online computational cost. 

The paper is organized as follows. Section \ref{sec:setup} introduces the setup of the problem. Section \ref{sec:method}  briefly reviews the POD-NN method from \cite{hesthaven2018non}, 
then  introduces our proposed method - BiFi-NN and discuss the error contribution of the proposed method. In Section \ref{sec:example}, we  illustrate the effectiveness of the proposed BiFi-NN algorithm via  several numerical examples. We conclude the paper in Section \ref{sec:summary}. 

\section{Problem Setup}
\label{sec:setup}
For simplicity of presentation, we consider the following parameterized PDEs:
\begin{equation}
\left\{
\begin{aligned}
& \mathcal{L}u(x, z) = f, &\ \text{in}\ D, \\
&u = g, &\ \text{on}\ \partial D,
\end{aligned}
\right.
\label{setup-PDEs}
\end{equation}
where $D\subset \mathbb{R}^n $ is the physical domain and the parameter $z\in I_z\subset \mathbb{R}^d $ with $d \ge 1$ represents the either the physical parameters or uncertain parameters in the model. 

Assume that we have two different models for \eqref{setup-PDEs}  available:
a high-fidelity solution $u_h(x, z): D\times I_z \to \mathbb{V}_h $ and a low-fidelity solution $u_l(x, z): D\times I_z \to \mathbb{V}_l $, where $\mathbb{V}_h $ is the discrete approximation space for the high-fidelity solution 
and $\mathbb{V}_l $ is the approximation space for low-fidelity solution
. Typically, the dimension of $\mathbb{V}_h $ is large  in order to resolve the details of the system with  high accuracy, while $\mathbb{V}_l $ is parameterized with fewer degrees of freedom than the high-fidelity space, i.e., $1 \le \dim \mathbb{V}_l \ll \dim \mathbb{V}_h $. Consequently, the high-fidelity solutions are more accurate and expensive to simulate, while the low-fidelity solutions are cheap to simulate. Even  though the accuracy of low-fidelity solution is typically lower than the high-fidelity solution, it is still capable of capturing the important features of the underlying problem. 


In this paper, our goal is to   efficiently construct a nonintrusive reduced model that is able to produce an accurate approximation of high-fidelity solution $u_h(z)$ with affordable online computational cost, particularly for nonlinear problems. To achieve this, we shall take advantage of the predictive capability of the neural network and the data from the low-fidelity and high-fidelity models. 



 

\section{Method}
\label{sec:method}

To set the stage for the later discussion, we first  briefly review the POD-NN algorithm proposed in \cite{hesthaven2018non}, then discuss the   proposed bi-fidelity data-assisted neural network approximation - the BiFi-NN method and a modification of the POD-NN method as a reference solution. Finally, we shall discuss the error contributions.

\subsection{POD-NN}
\label{sec:POD-NN}
The POD-NN method proposed in \cite{hesthaven2018non} basically consists of the following steps:

\begin{enumerate}[(1)]

\item Select a subset of parameters $\Gamma_P = \{z_1, z_2, \hdots, z_P\}\subset I_z $. For each sample $z_i\in \Gamma_P$, perform the corresponding high-fidelity full-order simulation to generate the snapshot matrix 
$$S= [u_h(z_1), u_h(z_2), \hdots, u_h(z_P)],$$ 
then construct the high-fidelity POD reduced basis set $V_h$.

\item Select a subset of parameters $\Gamma = \{z_1, z_2, \hdots, z_M\}\subset I_z $, independent of $\Gamma_P$. For each sample $z_i \in \Gamma$, perform a high-fidelity full-order simulation to get the high-fidelity solution $u_h(z_i)$. 
Then compute the first $r$ POD coefficients $c_h(z_i) = [c_{h,1}(z_i),\cdots, c_{h,r}(z_i)]^\top$ on the reduced approximation space.

\item Construct and train a multi-layer perceptron with the training set collected in the previous step, where the input is the parameter $z$ and the output is the high-fidelity POD coefficient vector $c_h(z)$.


 \item At the online stage, for a new parameter value $z^* $,  evaluate the trained network to predict the corresponding reduced coefficients $\tilde{c}_h(z^*) $ and then compute the reduced  solution. 

\end{enumerate}
In the following, we shall discuss each step of the POD-NN algorithm in detail.
\subsubsection{The proper orthogonal decomposition}  
\label{POD} 
In this section, we shall briefly introduce the proper orthogonal decomposition (POD)  \cite{chatterjee2000introduction}, which is the building block of the POD-NN method.  
As one of the most widely used methods to generate reduced basis, 
the general idea of POD is to seek for a set of parameter-independent function basis for the low-dimensional representation of the full-order solution space $\mathbb{V}_h $. 
Assume  we have a collection of  high-fidelity solutions at a selected set of parameter points $\Gamma_P\ = \{z_1, z_2, \hdots, z_P\}$ available,  which is assumed to be large and rich enough to represent the parameter space. We concatenate the high-fidelity solutions as the following snapshot matrix: 
\begin{equation}
S_h = [u_h(z_1),\ u_h(z_2),\ \hdots,\ u_h(z_P)] \in \mathbb{R}^{N_h\times P}.
\end{equation}
Each column is a high-fidelity solution $u_h(z_i)$ at the parameter $z_i$, of length $N_h $, reflecting the degrees of freedom of the high-fidelity solution.  With this snapshot matrix $S_h$, a singular value decomposition is then performed to reveal the reduced space:
\begin{equation}
S_h = U_h\Sigma_h W^\top_h,
\end{equation}
where $U_h\in \mathbb{R}^{N_h\times P}$ and  $W_h\in \mathbb{R}^{P\times P}$ are orthogonal matrices. The diagonal matrix $\Sigma_h = \text{diag}(\sigma_1, \sigma_2, \hdots, \sigma_R,0,\hdots,0) \in \mathbb{R}^{P\times P}$ satisfies $\sigma_1 \ge \sigma_2 \ge \hdots \sigma_R >0$. $R$ is the rank of the snapshot matrix. 
The first $r (< R)$ columns of $U_h$ are chosen to form the basis of the reduced space $\mathbb{V}_{r} $, i.e.,
\begin{equation}
\mathbb{V}_{r} = \text{span}\{U_{h,1}, U_{h,2}, \hdots, U_{h,r}\}.
\end{equation}
 Under the context of reduced-order modeling, we assume $r\ll N_h$. Once the reduced basis set is available, the reduced representation of the high-fidelity snapshot $u_h(z_j)$ can be written in the following form: 
\begin{equation}
u_r(z_j) = \sum_{i=1}^{r}c_{h,i}(z_j)U_{h,i},
\end{equation}
where the high-fidelity POD  coefficient vector $c_h(z_j) = [c_{h,1}(z_j), \hdots, \ c_{h,r}(z_j)]^\top \in \mathbb{R}^r $  can be computed by  projecting the full-order snapshot $u_h(z_j)$ onto the reduced approximation space $\mathbb{V}_r$:
\begin{equation}
c_{h}(z_j) = V^\top_h u_h(z_j),
\end{equation}
where the basis matrix $V_h = [U_{h,1}, U_{h,2}, \hdots, U_{h,r}]$.

 We shall emphasize that  traditional projection-based reduced basis methods need to project the full-order equation into the reduced approximation space to recover the high-fidelity POD coefficient vector for a new given $z$. We refer interested readers  to  \cite{quarteroni2015reduced,hesthaven2016certified} for more details of the traditional projection-based reduced basis methods. In contrast, the basic idea of POD-NN is to construct a surrogate of high-fidelity POD coefficients by a neural network so that  the projection of the high-fidelity model is not needed  during the online stage.


\subsubsection{Neural Network} The second key ingredient of POD-NN is the neural network approximation. Neural network is a universal function approximation model with the capability to learn from any type of observed data, thus provides an alternative to traditional function approximation methods \cite{schalkoff1997artificial}. One widely-used class of neural networks is the feedforward network, which is also called multi-layer perceptrons (MLPs) \cite{hornik1991approximation}. It consists of a collection of layers, including an input layer, an output layer, and a number of hidden layers. Each hidden layer contains a certain number of neurons, called hidden units, and a nonlinear activation function. For traditional MLPs, the connection between two continuous layers is an affine function defined by a set of weights and biases. 

\begin{figure}[htbp]
\centering
\includegraphics[scale=0.3]{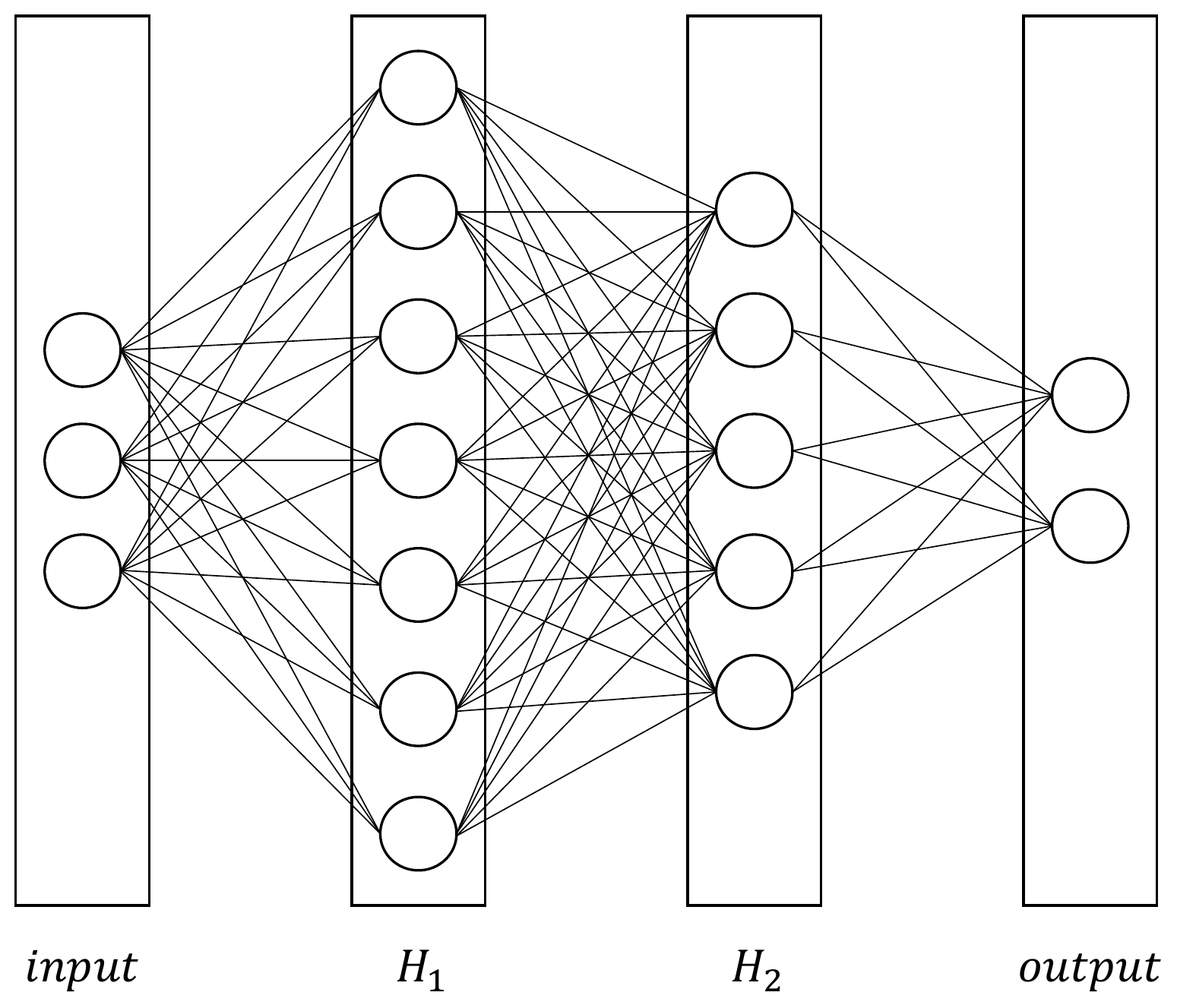}
\caption{The structure of a two-hidden-layer neural network with three input nodes and two output nodes.}
\label{structure-NN-nodes}
\end{figure}

Figure \ref{structure-NN-nodes} illustrates the structure of a two-hidden-layer MLP.
The circles represent the neurons in input, hidden and output layers, and information flows from left to right. 

As a crucial step to construct the neural network, the training of the neural network  generally involves gradient-based optimization. 
 The gradient information is usually computed through backpropagation. Widely used gradient-based training algorithms for large networks includes stochastic gradient descent, RMSprop, and Adam \cite{DBLP:journals/corr/Ruder16}. 
 
\subsubsection{POD-NN Algorithm}
\begin{figure}
    \centering
    \includegraphics[scale=0.8]{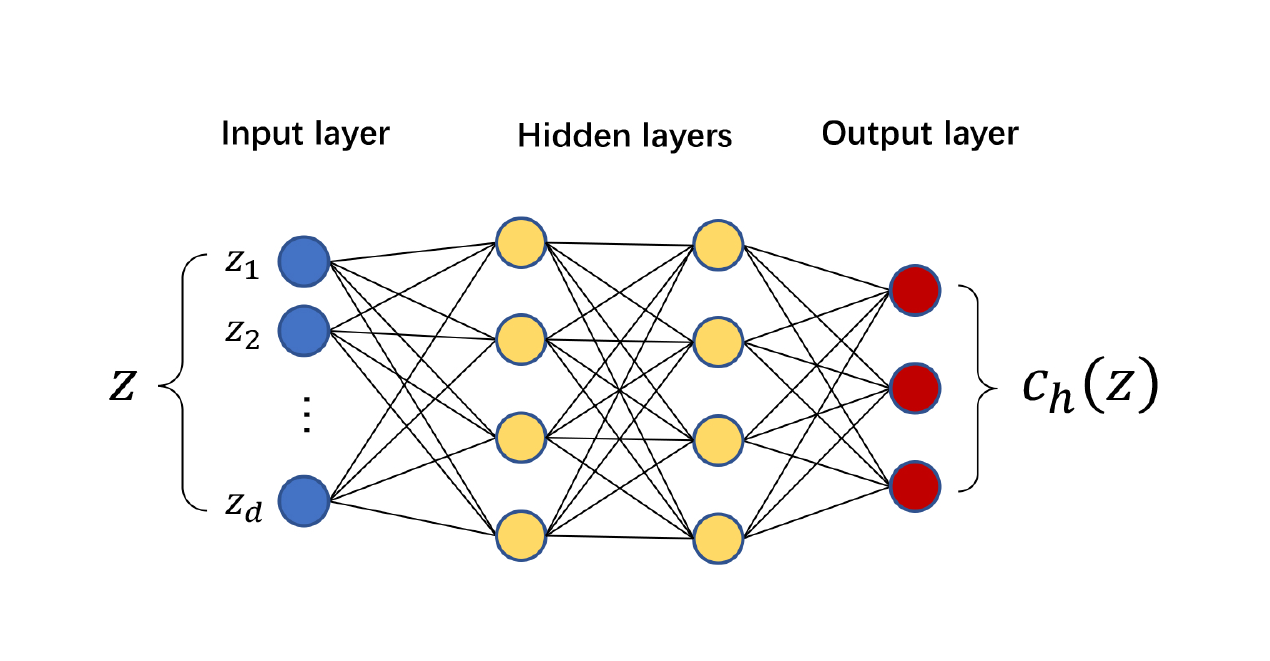}
    \caption{The network structure of the POD-NN method. The input is the parameter vector $z$ and the output is the high-fidelity POD coefficient vector $c_h(z)$.}
    \label{PODNN-structure}
\end{figure}

With the two building blocks discussed in the previous section, we discuss the basics of POD-NN in this section. POD-NN in \cite{hesthaven2018non} employed a shallow neural network to build a surrogate of high-fidelity POD coefficient. 
The input of the network is the parameter  $z$ and the output is the high-fidelity POD coefficient vector $c_h(z)$. It consists of two  dense hidden layers with the same number of hidden units, whose structure is illustrated in Figure \ref{PODNN-structure}. 

The resulted POD-NN approximation of the  high-fidelity POD coefficient vector $c_{h}(z)$  can be represented as follows:
\begin{equation}
\tilde{c}_h(z)= \Phi(z;\theta),
\end{equation}
where $\Phi(z;\theta)$ is the trained neural network approximation for the coefficient $c_{h}(z)$ and $\theta$ is the parameter of the neural network. 

Once the neural network is trained, we perform a forward pass to  predict the reduced coefficient $\tilde{c}_h(z^*)$ for a new given $z^*$, during the online stage. The corresponding reduced solution is given by 
\begin{equation}
\label{eq:redsol_podnn}
\tilde{u}_h(z^*) = V_h\tilde{c}_h(z^*), 
\end{equation}
where $V_h$ is the high-fidelity POD basis set. We emphasize  that the key advantage of this approach is that the high-fidelity model is completely decoupled from the online stage.  

This completes the description of the original POD-NN algorithm. The detailed steps of the corresponding offline and online algorithms are summarized in Algorithm \ref{POD-NN Offline} and Algorithm \ref{POD-NN Online}. We refer interested readers  to \cite{hesthaven2018non} for more details of POD-NN. 

\begin{algorithm}[h]
\caption{Offline Stage for POD-NN}
\label{POD-NN Offline}

Sample a collection of parameters $\Gamma\ = \{z_1, z_2, \hdots, z_M\}\subset I_z $.

Run the high-fidelity model for each $z_j \in \Gamma $.


Compute the POD coefficient vector $c_h(z_j)$ for each high-fidelity snapshot $u_h(z_j)$  by projection:
$$
c_h(z_j) = V_h^\top u_h(z_j),\quad j=1,\hdots, M.
$$


Train a network $\Phi(z;\theta)$ with the input $z$ and the output $c_{h}(z)$.
\end{algorithm}

\begin{algorithm}[h]
\caption{Online Stage for POD-NN}
\label{POD-NN Online}


Evaluate the trained network $\Phi(z; \theta)$ at the given parameter $z^*$ to predict  the high-fidelity POD coefficient  $\tilde{c}_{h}(z^*)$

Compute the POD-NN approximation of the high-fidelity solution for the given $z^*$:
$$
\tilde{u}_h(z^*) = V_h\tilde{c}_h(z^*).
$$

\end{algorithm}

\subsection{Bi-Fidelity Data-Assisted Neural Network} As we discussed in the previous section, POD-NN focuses on learning the map from the parameter space $I_z $ to the high-fidelity POD coefficient space $C_h \in \mathbb{R}^r$. Nevertheless, from the view of the conventional wisdom in the machine learning community,  the parameter $z$ as the input feature is not data-independent and might not be   strongly informative nor encode  much prior knowledge. 

Alternatively, since low-fidelity models are cheap to evaluate and can capture some important information of the underlying physical systems, it would be desirable to use low-fidelity data to extract useful features to improve the predictive performance of the neural network approximation. In this paper, we propose a novel bi-fidelity data-assisted neural network approximation (BiFi-NN) by modifying  the original POD-NN algorithm: during the offline stage, we first employed the  POD coefficient $c_l(z)$ of the low-fidelity model as the augmented data-dependent features, and train the neural network to predict high-fidelity POD coefficient $c_h(z)$. In contrast to the original POD-NN, the learned mapping  is now from the combined feature space (the original parameter space $I_z$ and the low-fidelity coefficient space $C_l  \in \mathbb{R}^r)$ to the  high-fidelity POD coefficients space $C_h$. During the online stage, it only requires one cheap low-fidelity simulation run to generate the low-fidelity POD coefficients to predict the high-fidelity POD coefficients from the pre-trained neural network. Such an approach not only allows us to incorporate more relevant features to improve the predictive performance, but also remove the dependence on the high-fidelity solver from the online stage. 

In the following, we shall present the details of the offline and online stage of the proposed BiFi-NN algorithm. 
 
\subsubsection{Offline Stage}
\label{offline-1}
 There are two major steps at offline stage: (1) prepare the data. (2) train the neural network. We shall detail each step in this section.


{\bf{Data Preparation}}. 
We first sample a collection of parameters $\Gamma_P =\{z_1, z_2, \hdots, z_P\}$ $\subset I_z $ and run the corresponding low-fidelity and high-fidelity simulations. Based on the acquired simulation data, we compute  the corresponding POD basis set $V_l $ and $V_h$. To prepare the training input-output pairs for the next step,  we  run low-fidelity and high-fidelity simulations over a sample set $\Gamma =\{z_1, z_2, \hdots, z_M\}\subset I_z $, independent of $\Gamma_P$. And then we compute the corresponding low-fidelity and high-fidelity POD coefficients: $c_l(z_j) $, $c_h(z_j)$, $\forall z_j\in \Gamma$.

\begin{figure}[htbp]
\centering
\includegraphics[scale=0.8]{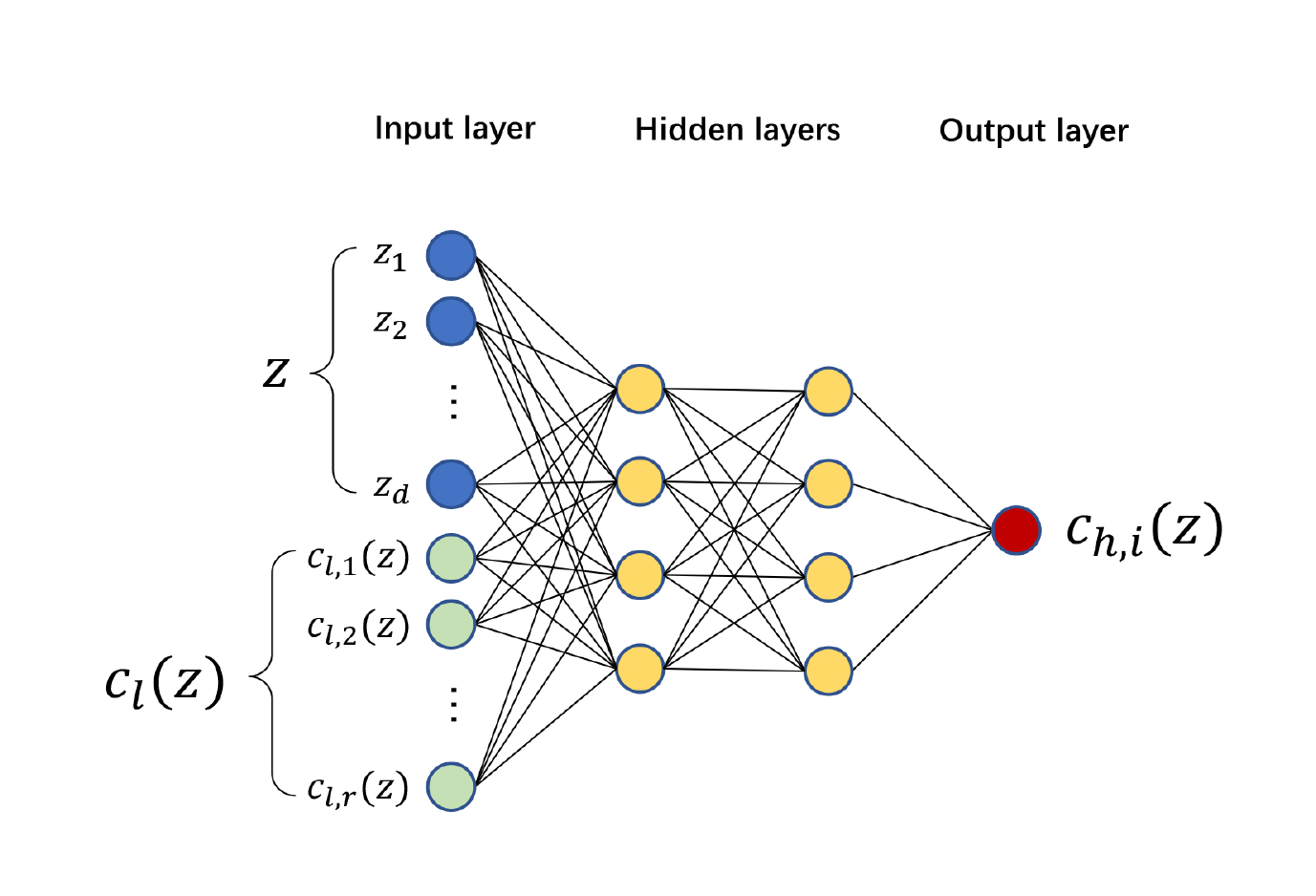}
\caption{The network structure of the $i^{th}$ net of the proposed BiFi-NN method. The input is the concatenation of the parameter vector $z$ and the low-fidelity POD coefficient $c_l(z)$, and the output is the $i^{th}$ component of  high-fidelity POD coefficient $c_{h}(z)$.}
\label{BiFi-structure}
\end{figure}

{\bf{Architecture and Training of BiFi-NN}}. 
 Once the data is available, we shall construct a surrogate for the high-fidelity POD coefficients. 
More specifically, we shall train $r$ neural networks to predict the $r$ components of the high-fidelity coefficient $c_h(z)$ separately. 
The structure of the $i^{th}$ network consisting of two hidden layers is illustrated in Figure \ref{BiFi-structure}, where the numbers of hidden units in both layers are equal, and the training input is given by  concatenating the parameter $z$ and the low-fidelity POD coefficient vector $c_l(z)$ together, i.e.,
\begin{equation}
x = (z, c_l(z)) \in I_z\times C_l,
\end{equation}
and the output $y$ is the $i^{th}$ component $c_{h,i}(z)$ of the high-fidelity POD coefficient $c_{h}(z)$ We remark that the structure of BiFi-NN is similar with POD-NN. However, there are two major differences between BiFi-NN and POD-NN: (1) We incorporate additional new features from the low-fidelity data in order to improve the predictive power of the neural network (2) Instead of training a single neural network to predict all components of the high-fidelity coefficient $c_h(z)$ (referred to as the joint approach), we train $r$ neural networks to predict each component of the high-fidelity coefficient $c_h(z)$ separately,  whose  predictive accuracy should be comparable to the joint approach. Nevertheless, BiFi-NN is  expected to be relatively easy to train and more memory-efficient due to the compact configuration (single output node for each network).   

Consequently, the BiFi-NN approximation of the $i^{th}$ component of the high-fidelity POD coefficient $c_{h}(z)$  can be represented as follows:
\begin{equation}
\tilde{c}_{h,i}(z)= \Phi_i(x;\theta), \quad i = 1,\hdots, r,
\end{equation}
where $\Phi_i$ is the trained neural network approximation for the $i^{th}$ component of the high-fidelity coefficient $c_{h}(z)$ and $\theta$ is the parameter of the network.


We remark that the high-fidelity models are expensive to simulate, hence the number of high-fidelity training data is limited. 
To avoid over-fitting, we limit ourselves to a two-hidden-layer structure which theoretically can  approximate any function \cite{university1988continuous,cybenko1989approximation}. In addition, since the neural network is shallow, there is  less concern about the vanishing of gradients. Therefore, we choose to use \texttt{tanh} as the activation function to make use of its nonlinearity. 
To train the neural network, we  optimize the mean squared loss function by the Levenberg-Marquardt (LM) algorithm, which is suitable to train a shallow network with a small number of connections as suggested in \cite{hagan1994training,hesthaven2018non}. 


Algorithm \ref{BiFi-NN Offline} outlines the detailed steps for the offline stage.

\begin{algorithm}[ht]
\caption{Offline Stage for BiFi-NN}
\label{BiFi-NN Offline}

Sample a collection of parameters $\Gamma = \{z_1, z_2, \hdots, z_M\}\subset I_z $.

Run the low-fidelity model $u_l(z_j)$ and high-fidelity model $u_h(z_j)$ for each $z_j \in \Gamma$.



Compute the POD coefficients for both fidelities by  projection:
$$
c_l(z_j) = V_l^\top u_l(z_j), \ c_h(z_j) = V_h^\top u_h(z_j), \quad j=1,\hdots, M.
$$

Concatenate the parameter $z_j$ with the corresponding low-fidelity coefficient $c_l(z_j)$ to form the input for the proposed neural network:
$$
x_j = (z_j, c_l(z_j)).
$$

For $i = 1, \hdots, r$, train the network $\Phi_i(x; \theta)$ with $c_{h,i}(z)$ as the output independently.
\end{algorithm}

\subsubsection{Online stage} 
Once the neural network is trained, we can predict the high-fidelity POD projection coefficients very efficiently. 
Given a new parameter $z^* $, one needs to 
\begin{itemize}
\item Run the low-fidelity model and compute its $r$ low-fidelity POD coefficients via a projection onto the low-fidelity reduced approximation space, i.e.,
\begin{equation}
c_l(z^*) = V_l^\top u_l(z^*).
\end{equation}
and concatenate the learned low-fidelity coefficients with $z^* $:
\begin{equation}
\ x^* = (z^*, c_l(z^*)).
\end{equation}
\item For $i = 1, \hdots, r$, evaluate the pre-trained network $\Phi_i (x^*; \theta)$ to approximate the $i^{th}$ component of the corresponding high-fidelity POD coefficients $c_{h}(z^*)$, and the results are concatenated to obtain the POD coefficient vector, i.e.,
\begin{equation}
\tilde{c}_{h,i}(z^*) = \Phi_i(x^*;\theta), \quad i=1,\cdots,r.
\end{equation}

\item The resulted BiFi-NN approximation of the high-fidelity solution for the given parameter $z^*$ is given by
\begin{equation}
\tilde{u}_h(z^*) = V_h\tilde{c}_h(z^*).
\end{equation}
\end{itemize}

We emphasize that for a new given parameter $z^*$, our method only requires one additional low-fidelity  run to extract the additional input feature 
to predict the high-fidelity reduced coefficient during the online stage. Therefore, the online cost mainly depends on the cost of low-fidelity solvers. Since we assume the low-fidelity model is cheap to compute, this cost should be affordable. 

Algorithm \ref{BiFi-NN Online} outlines the details for the online stage.

\begin{algorithm}[h]
\caption{Online Algorithm of BiFi-NN}
\label{BiFi-NN Online}

Run the low-fidelity model $u_l(z^*) $ for the given $z^* $.

Compute the low-fidelity POD coefficients and concatenate the learned low-fidelity coefficients with $z^* $: 
$$
c_l(z^*) = V_l^\top u_l(z^*), \ x^* = (z^*, c_l(z^*)).
$$

For $i = 1, \hdots, r$, evaluate the trained neural network $\Phi_i$ at the combined feature $x^*$ to the BiFi-NN approximation of the $i^{th}$ component of the high-fidelity POD coefficient:
$$
\tilde{c}_{h,i}(z^*) = \Phi_i(x^*;\theta), \quad i=1,\cdots, r.
$$
and concatenate the results to obtain the coefficient vector $\tilde{c}_h(z^*)$.

Compute the BiFi-NN approximation of the high-fidelity solution $u_h(z^*)$ for the given $z^*$:
$$
\tilde{u}_{h}(z^*) = V_h\tilde{c}_h(z^*).
$$
\end{algorithm}


\subsection{A modified POD-NN Algorithm} \label{Modified POD-NN}

\begin{figure}
    \centering
    \includegraphics[scale=0.8]{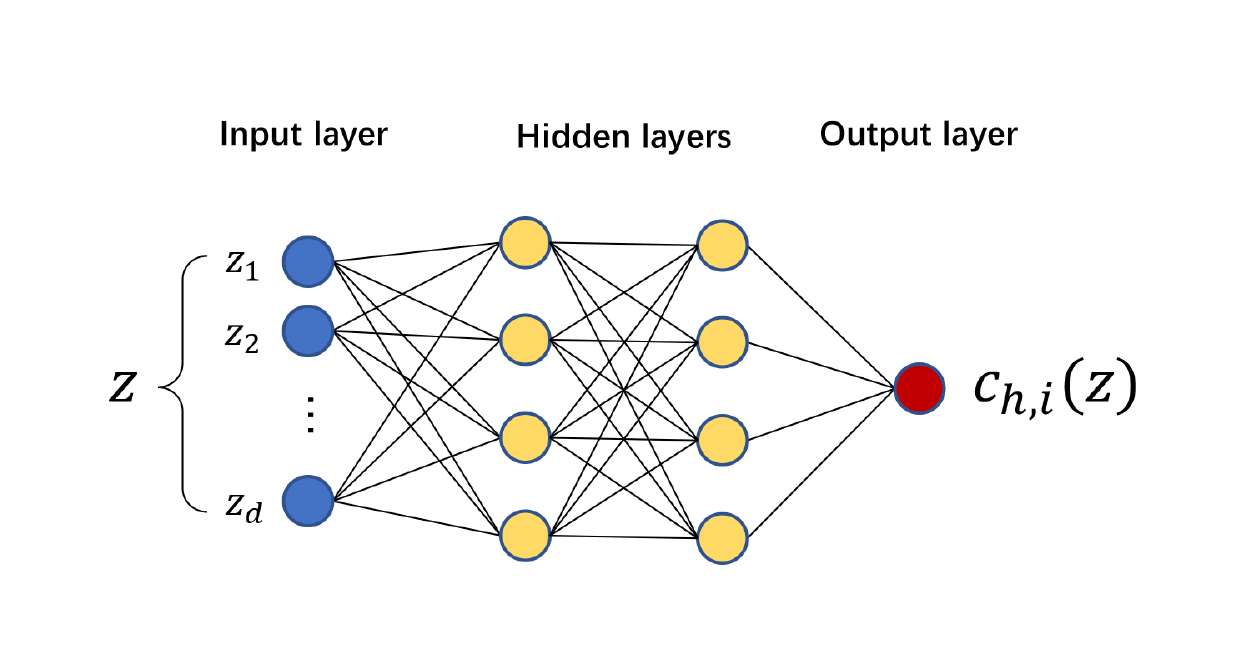}
    \caption{Network structure of the $i^{th}$ net for modified POD-NN method. The input is the parameter vector $z$, and the output is the $i^{th}$ component of the high-fidelity POD coefficient $c_{h}(z)$.}
    \label{MPODNN-structure}
\end{figure}

From the discussion in the previous section, BiFi-NN trains $r$ neural networks to predict the high-fidelity POD coefficient $c_h(z)$, where the output for the $i^{th}$ net is the $i^{th}$ component of the high-fidelity coefficient $c_{h}(z)$. 
In order to demonstrate the effectiveness of the additional input features of BiFi-NN,  we proposed to slightly modify the POD-NN structure so that it has the same output layer with the BiFi-NN and the rest of architecture is the same with that of the original POD-NN discussed in Section~\ref{sec:POD-NN}. 
The structure of the $i^{th}$ net is illustrated in Figure \ref{MPODNN-structure}. All $r$ nets have the same structure for the sake of simplicity. We referred this method as {\it{modified POD-NN}} (MPOD-NN).

Consequently, the modified POD-NN approximation of the $i^{th}$ component of the high-fidelity POD coefficient vector $c_{h}(z)$  can be represented as follows:
\begin{equation}
\tilde{c}_{h,i}(z)= \Phi_i(z;\theta), \quad i = 1,\hdots, r,
\end{equation}
where $\Phi_i(z;\theta)$ is the trained neural network approximation for the coefficient $c_{h, i}(z)$  and $\theta$ is the parameter of the network. 

During the online stage, for each given parameter $z^*$, we obtain the $i^{th}$ component of high-fidelity POD coefficient $\tilde{c}_{h}(z^*)$ by a forward pass through the $i^{th}$ pre-trained net. All these predicted results are concatenated into the POD coefficient vector $\tilde{c}_h(z^*) = [\tilde{c}_{h,1}(z^*),  \hdots, \tilde{c}_{h,r}(z^*)]^\top$. The corresponding reduced solution of the high-fidelity solution $u_h(z^*)$ is then given by
\begin{equation}
    \tilde{u}_h(z^*) = V_h\tilde{c}_h(z^*).
\end{equation}
 For the same amount of training samples, the predictive accuracy of the modified POD-NN is expected to be comparable with the original POD-NN. In contrast to $r$ output nodes in the original POD-NN,  each net in modified  POD-NN only has a single output node. Therefore  each net in the modified POD-NN is is relatively easy to train with the Levenberg-Marquardt algorithm due to the compactness of the network.


\subsection{Discussion of Error Contributions}
\label{sec:err_contr}
Since the error estimation for neural network approximation is  still under development in the literature, a detailed error analysis is difficult to perform.  Here, we briefly discuss the error contribution of the approximation for the high-fidelity solution. To estimate the error, it is sufficient to consider the approximation error between the high-fidelity solution and the reduced solution at each parameter point $z$:
\begin{equation}
\label{error-realizatin}
\begin{aligned}
 e_a(z) = \frac{\lv u_h(z) -  \tilde{u}_h(z)\rv}{\lv u_h(z) \rv} 
 &= \frac{\lv u_h(z) - V_hc_h(z)+V_h(c_h(z)-\tilde{c}_h(z))\rv}{\lv u_h(z) \rv} \\
&= \frac{\lv u_h(z) - V_hV_h^\top u_h(z)+V_h(c_h(z) - \tilde{c}_h(z))\rv}{\lv u_h(z) \rv} \\
&\le \frac{\lv u_h(z) - V_hV_h^\top u_h(z)\rv+ \lv V_h(c_h(z) - \tilde{c}_h(z))\rv}{\lv u_h(z) \rv} \\
& \le e_{p}(z) + \lv V_h\rv\frac{\lv c_h(z)-\tilde{c}_h(z)\rv}{\lv u_h(z)\rv}\\
&\le e_{p}(z) + e_c(z),
\end{aligned}
\end{equation}
where $\lv \cdot \rv$ is the Euclidean norm in $\mathbb{R}^{N_h} $. To get the last inequality, we used the fact that the POD basis $V_h$ is orthonormal. $e_a$, $e_p$ and $e_c$ are defined as follows: 
\begin{itemize}
\item $e_a(z)$ represents the approximation error  of the reduced solution $\tilde{u}_{h}(z)$, which is measured by the relative $L_2$ error between the reduced solution and the high-fidelity solution:
\begin{equation}
e_{a}(z) = \frac{\lv u_h(z) - \tilde{u}_h(z)\rv}{\lv u_h(z)\rv} = \frac{\lv u_h(z) - V_h\tilde{c}_h(z)\rv}{\lv u_h(z)\rv}.
\end{equation}

\item $e_p(z)$ represents the projection error of the high-fidelity solution $u_h(z)$, which is measured by the relative $L_2$ error between  the high-fidelity solution $u_h(z)$ and its projection onto the high-fidelity reduced approximation space $\mathbb{V}_r$:
\begin{equation}
e_{p}(z) = \frac{\lv u_h(z) - V_hV_h^\top u_h(z)\rv}{\lv u_h(z)\rv}.
\end{equation}
Assume a reasonable good reduced approximation space exists, the projection error can be reduced by increasing the dimension of the reduced approximation space.

\item $e_c(z)$ represents the coefficient error due to the neural network approximation, which is measured by the relative $L_2$ error between the learned POD coefficient vector $\tilde{c}_h(z)$ and the high-fidelity coefficient vector $c_h(z)$:
\begin{equation}
\label{coeff_err}
e_{c}(z) = \frac{\lv c_h(z)-\tilde{c}_h(z)\rv}{\lv u_h(z)\rv}.
\end{equation}
We remark that the error $e_c$ depends many factors, such as the optimization algorithms for training, choice of loss functions, choice of the input features and available training points. In this project, we mainly explore the role of the features  extracted from the low-fidelity data on the predictive accuracy of the proposed neural network.
 
\end{itemize}

On the other hand, the projection error is the smallest distance between $u_h $ and the reduced space $\mathbb{V}_r$, i.e.,
\begin{equation}
e_p(z) = \min_{\tilde{u}_h\in\mathbb{V}_r} \frac{\lv u_h(z) - \tilde{u}_h(z)\rv}{\lv u_h(z)\rv},
\end{equation}
hence it provides a lower bound for the  approximation error $e_a $. Consequently, we have
\begin{equation}
\label{error-relation-end}
e_{p}(z)\le e_{a}(z) \le e_p(z) + e_c(z).
\end{equation}

The inequalities \eqref{error-relation-end} gives both upper and lower bound for the  approximation error of the reduced solution expressed in terms of the projection error and the coefficient error. This also reveals the approximation error $e_a$'s dependence on two major error contributions - the coefficient error committed by neural network approximation and the projection error of high-fidelity solutions on the reduced space.  Even though this is a rough analysis of the error contribution, it still provide a general guideline to analyze the error behavior of the proposed method in the next section.

\begin{remark}
By the linearity of the expectation operator and \eqref{error-relation-end}, we can also get the error bound of the   mean relative approximation error as follows:
 \begin{equation}
\label{error-mean}
\mathbb{E}[e_{p}(z)]\le \mathbb{E}[e_{a}(z)] \le \mathbb{E}[e_p(z)] + \mathbb{E}[e_c(z)].
\end{equation}
\end{remark}

\section{Numerical Examples}
\label{sec:example}
In this section, we present several numerical examples to illustrate the effectiveness and performance of the proposed method. To measure the accuracy of the approximation, we shall compute the following three types of errors over an independent test of size $M$: 


\begin{enumerate}[(a)]

\item The mean approximation error of the reduced solution $\tilde{u}_h(z)$, measured by the relative $L_2$ error with respect to the high-fidelity solution $u_h(z)$ :
\begin{equation}
\varepsilon_a = \frac{1}{M} \sum_{i=1}^M \frac{\lv u_h(z_i) - \tilde{u}_h(z_i)\rv}{\lv u_h(z_i)\rv},
\end{equation}
where $\lv \cdot \rv$ is the Euclidean norm in $\mathbb{R}^{N_h} $.


\item The  coefficient learning error for the first $r$ POD coefficients, measured by the mean relative $L_2$ error
\begin{equation}
\varepsilon_{c} = \frac{1}{M} \sum_{i=1}^M\frac{\lv c_{h}(z_i)-\tilde{c}_{h}(z_i)\rv}{\lVert u_h(z_i)\rVert},
\end{equation}
where $c_{h}(z) = [c_{h,1}(z),\hdots,c_{h,r}(z)]^\top$ is the coefficient vector of the first $r$ POD coefficients.
\item The mean relative POD projection error for high-fidelity solution,
\begin{equation}
\varepsilon_p = \frac{1}{M} \sum_{i=1}^M \frac{\lv u_h(z_i) - V_hV_h^\top u_h(z_i)\rv}{\lv u_h(z_i)\rv}.
\end{equation}

\end{enumerate}

By the similar procedure in  Section \ref{sec:err_contr}, we can derive a similar result: 
\begin{equation} \label{error-bound}
    \varepsilon_p \le \varepsilon_a \le \varepsilon_p+\varepsilon_{c}.
\end{equation}
The inequality \eqref{error-bound} is a discrete version of \eqref{error-mean}.


For all examples in the rest of the section, we employ a two-hidden-layer neural network for both modified POD-NN and BiFi-NN.
We remark that an additional dataset is used for validation, whose size is 25\% of the training set.
 To find the best network configuration, i.e, the number of hidden units $H$, we choose the best results for the modified POD-NN and BiFi-NN  over the number of hidden units $H$ (varying from 1 to 24). For simiplicity, we set two hidden layers with the same number of hiddent units $H_1=H_2=H$. The optimization is carried out by the  Levenberg-Marquardt (LM) algorithm as mentioned before.  The multiple restarts approach is employed to prevent the results from depending on
the way the weights are (randomly) initialized. In other words, for each configuration, we train 10 nets with random initial conditions, and select the network with the smallest validation error.  

With loss of generality, we employed the solutions solved on coarse and fine meshes as the  low-fidelity and high-fidelity models in our numerical examples due to their availability for most applications. Nevertheless, the method itself has no restriction on both models employed if they model the same physical system. 
\subsection{1D stochastic elliptic equation} 
We first consider a 1D elliptic equation with random diffusivity coefficient, a standard benchmark problem in the context of uncertainty quantification as follows: 
\begin{equation}\label{example4}
\left\{ 
\begin{aligned}
-&(a(x, z)u'(x))' = 1, \quad x\in (0, 1), \\ 
&u(0) = u(1) = 0,
\end{aligned}
\right.
\end{equation}
with the random diffusivity coefficient $a(x, z)$ given as follows:
\begin{equation}
a(x, z) = 1+\frac{1}{2}\sum_{k=1}^{d}\frac{1}{k\pi}\cos(2k\pi x)z_k.
\end{equation}
The parameter $z = (z_1, z_2, \cdots, z_d)$, where each coordinate of $z$ is a uniformly distributed random variable in $[-1, 1]$. We fix the dimension $d = 10$ and therefore, it is a 10-dimensional problem in the parameter space.

We solve \eqref{example4} by  Chebyshev collocation method (in physical space).  We employ $K_h=128$ Chebyshev collocation points for the high-fidelity model and $K_l=32$ collocation points for low-fidelity models. In all cases, the models are evaluated on a 100-point uniform mesh in the physical space. The error metrics are evaluated on a  test set of 100 Monte Carlo points, and the reduced basis sets are computed with 100 snapshot solutions independent from both training and test sets. In the following tests, we shall train both BiFi-NN and modified POD-NN with the same training sets of different sizes and present the results of each method with the corresponding optimal hidden units ($H_1=H_2=H$) over the same test set. 

\begin{figure}[htbp]
\centering


\vbox{
\includegraphics[scale=0.42]{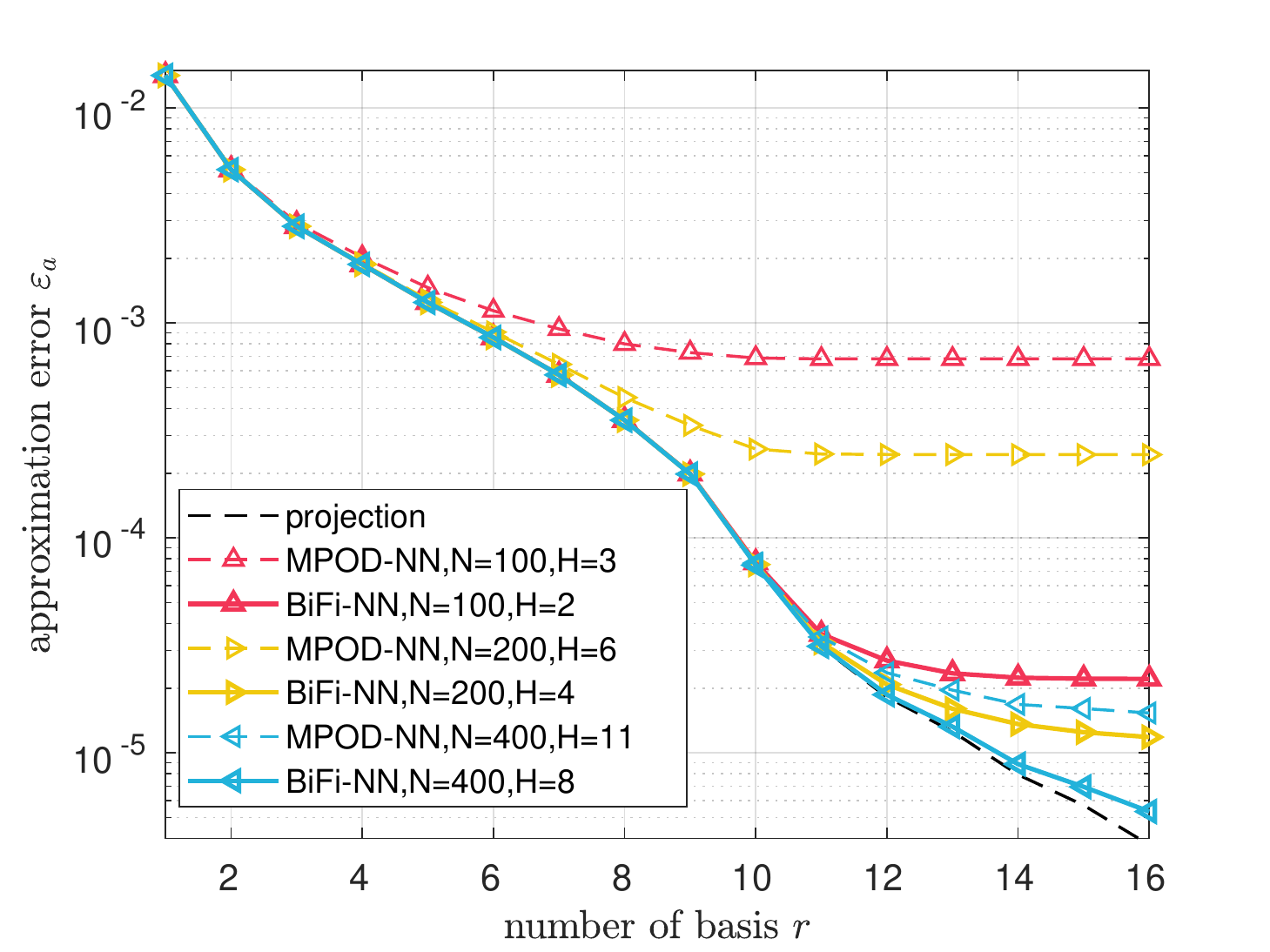}
\includegraphics[scale=0.42]{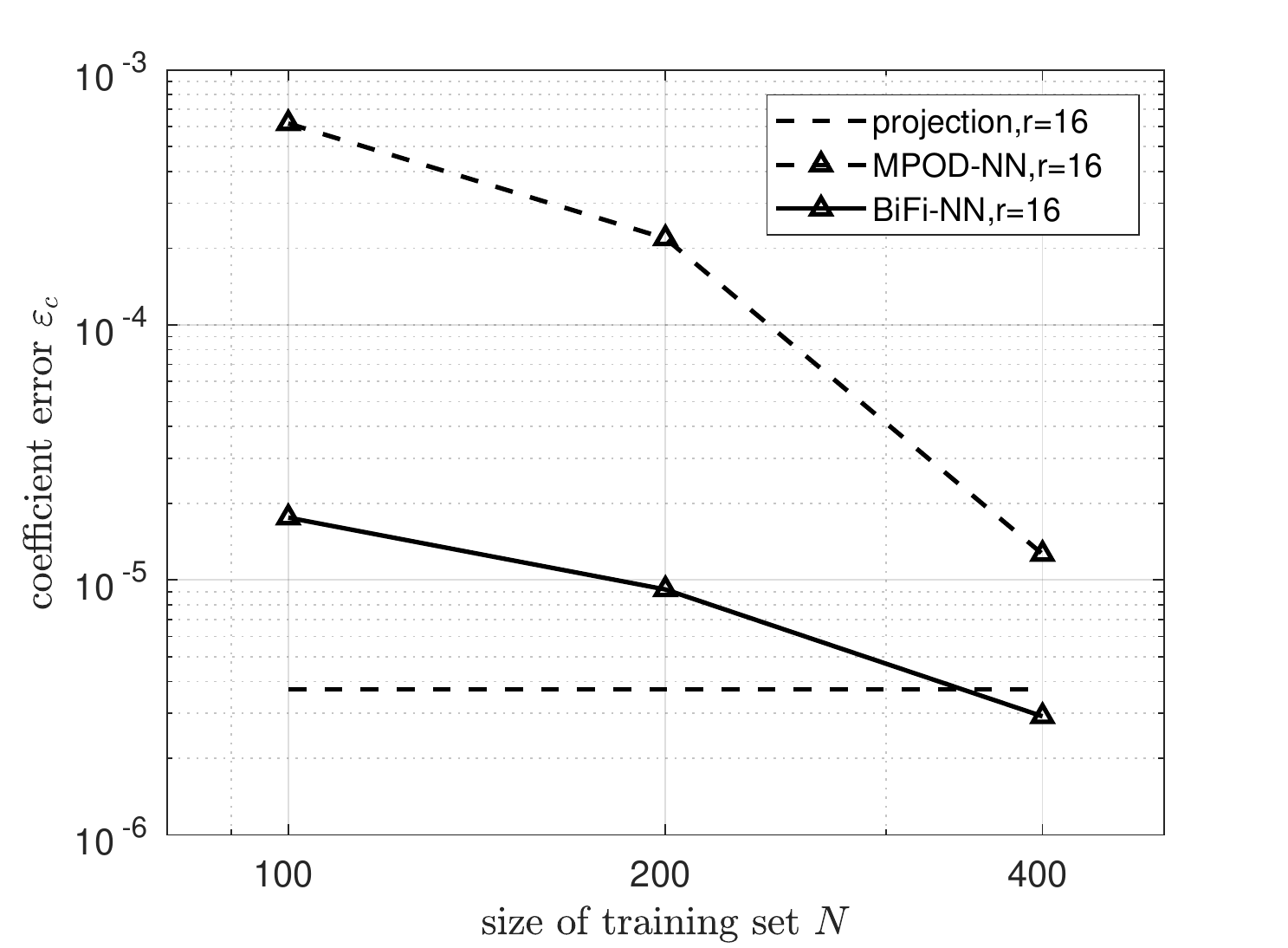}
}
\caption{Left: Convergence results of approximation error $\varepsilon_a$ by  modified POD-NN and BiFi-NN (based on the low-fidelity model \eqref{example4} with $K_l=32$) with the training sets of different sizes ($N=100,200,400$). Right: Coefficient error with $r = 16$ by modified POD-NN and BiFi-NN with respect to the size of training set compared to the projection error. The solid lines are results of BiFi-NN, and the dashed lines represent modified POD-NN.}
\label{example4-errors}
\end{figure}


Figure \ref{example4-errors} (left) shows the approximation error of modified POD-NN and BiFi-NN  as well as the projection error 
with respect to the number of  the high-fidelity POD basis. Both modified POD-NN and BiFi-NN continue to decrease as the number of POD basis increases and saturate later on. The saturation is because the coefficient error is dominant over the projection error when the dimension of the reduced approximation space is not  large enough.  When more training data is available, the saturation level can be further deduced by increasing the size of the network  
shown in Figure \ref{example4-errors} (left).  Moreover, BiFi-NN can further saturate at a much lower level than modified POD-NN for a fixed training set. 
This indicates that the additional input features we incorporated in BiFi-NN can effectively improve the predictive capability of the neural network. 


To better understand the properties of BiFi-NN, we plot the coefficient error by modified POD-NN and BiFi-NN for a fixed reduced dimension $r=16$ with respect to the size of training set  shown in Figure \ref{example4-errors} (right). It is clear that the coefficient error begins to dominate when the training set is small.  As we increase the number of the training set, both modified POD-NN and BiFi-NN get more accurate results. It is evident that BiFi-NN achieves a smaller coefficient error compared to modified POD-NN,  due to the effectiveness of  the low-fidelity feature. 

We remark that this is a linear problem, therefore traditional projection-based reduced basis method can also produce accurate results via a reasonable number of high-fidelity reduced basis. We present this example only for benchmark purposes to examine the accuracy of the proposed method. 

\subsection{2D Nonlinear Elliptic equation}
We next consider the following parameterized 2D nonlinear elliptic equation \cite{chaturantabut2010nonlinear} to illustrate its performance on nonlinear PDEs :
\begin{equation}
\label{example9}
    -\Delta u(x, y) +s(u(x, y); \mu) = 100\sin(2\pi x)\sin(2\pi y),
\end{equation}
where
\begin{equation}
    s(u; \mu) = \frac{\mu_1}{\mu_2}(e^{\mu_2 u}-1),
\end{equation}
 with a homogeneous Dirichlet boundary condition. The spatial domain is $(x, y) \in \Omega = (0, 1)^2$, The parameters are $\mu = (\mu_1, \mu_2) \in [0.01, 10]^2$. 

We solve this problem using $P_1$ finite element. For low-fidelity solutions, we use 135 elements, while for high-fidelity model, we use 2960  elements. 
The training set is sampled by Latin hypercube sampling (LHS) \cite{iman2014atin}, while the test set is generated on 256 uniform grids in the parameter space, and the reduced basis set is  generated from 225 uniform grid points in the parameter space.

\begin{figure}[htbp]
\centering
\vbox{
\includegraphics[scale=0.42]{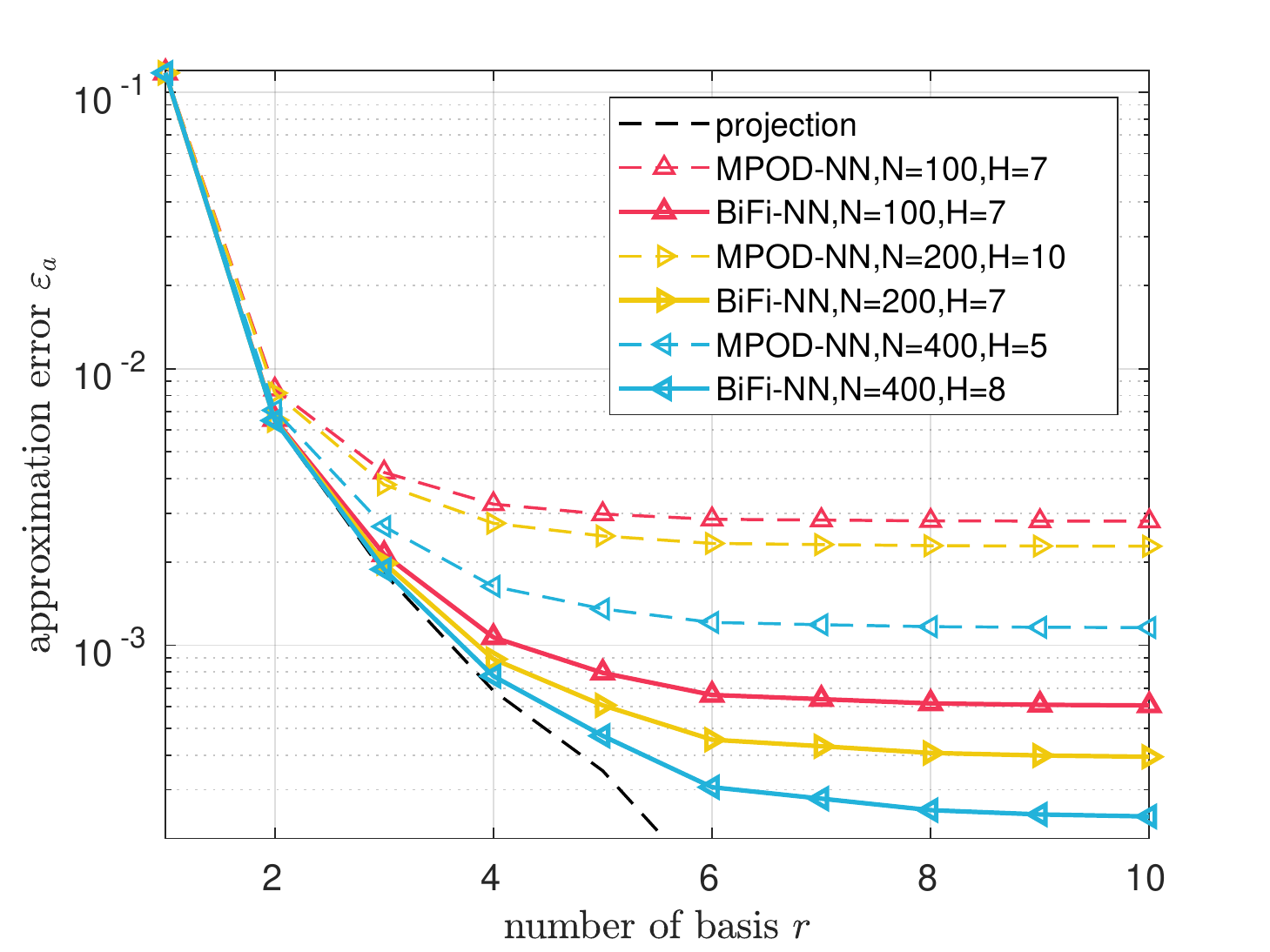}
\includegraphics[scale=0.42]{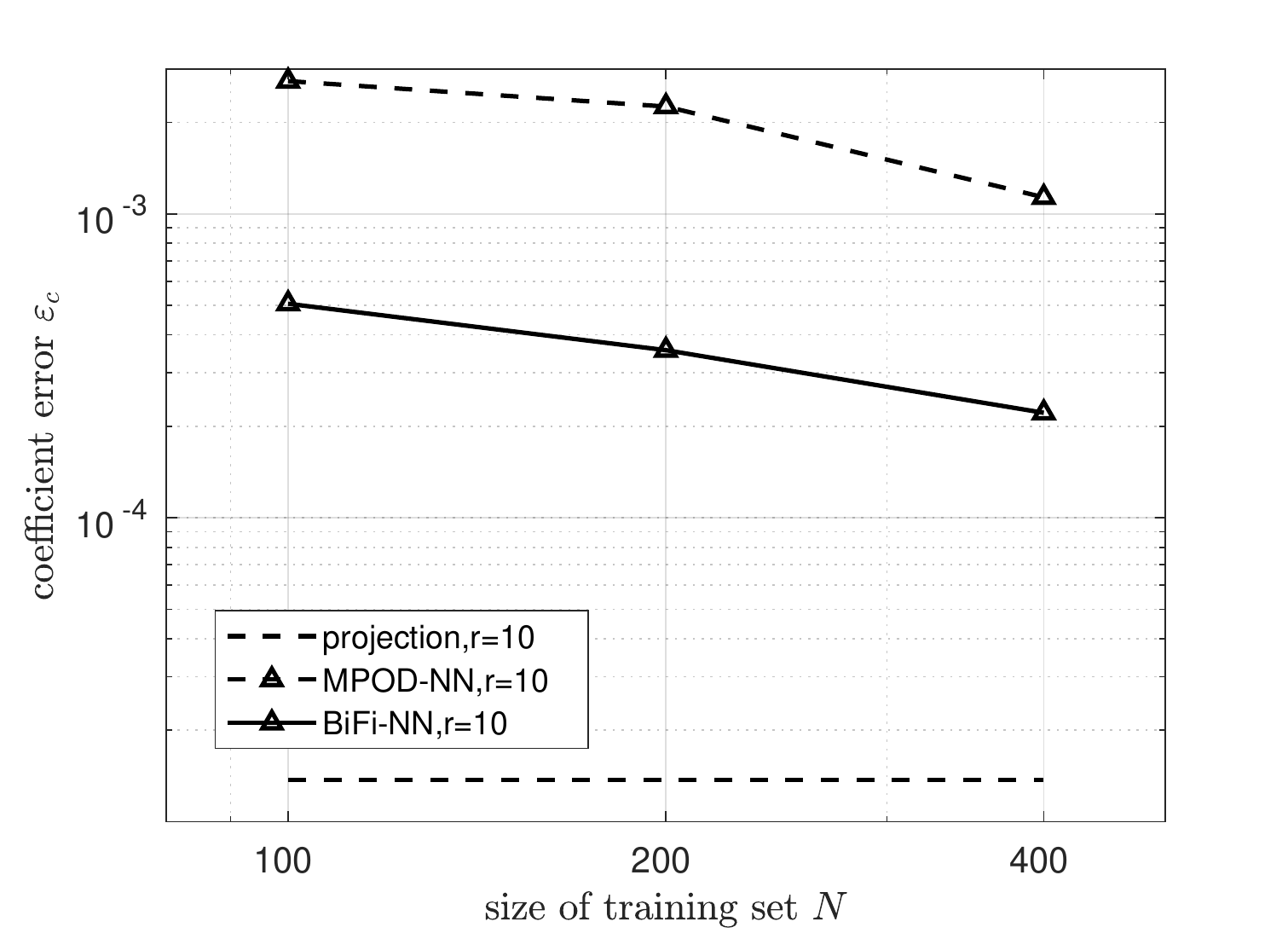}
}
\caption{Left: Convergence results of approximation error $\varepsilon_a$ by modified POD-NN and BiFi-NN for problem \eqref{example9} with training sets of different sizes ($N=100,200,400$). Right: Coefficient error with $r = 10$ by modified POD-NN and BiFi-NN with respect to the size of training set compared to the projection error. The solid lines are results of BiFi-NN, and the dashed lines represent modified POD-NN.}
\label{example9-errors}
\end{figure}

The approximation error of modified POD-NN and BiFi-NN are plotted in Figure~\ref{example9-errors} (left). We first observed that both modified POD-NN and BiFi-NN enjoy fast decay with respect to the number of high-fidelity POD basis and resemble the project error. As the number of high-fidelity basis increases, modified POD-NN saturates quickly and while BiFi-NN can continue to decrease and saturates at a lower level, indicating the effectiveness of low-fidelity features. 
We also investigate the effects of the size of the training set and report the results  in Figure~\ref{example9-errors} (left). It is clear that the predictive accuracy of both modified POD-NN and BiFi-NN can be further improved when more training date is available.
Overall, BiFi-NN produced a lower coefficient error.

We next  fix the reduced dimension  $r = 10$, which results in a small projection error shown in Figure~\ref{example9-errors} (right). 
We expect to see the dominance of the coefficient errors due to neural network approximation. This is
clearly illustrated in Figure~\ref{example9-errors} (right), which plots the coefficient errors with
respect to the size of the training set. The coefficient errors are much larger
than the projection error in  Figure~\ref{example9-errors} (right) confirming that in this case, the largest error contribution
stems from the coefficient error. In addition, BiFi-NN has a similar convergence rate with modified POD-NN,
but the coefficient error is roughly one order smaller than that of modified POD-NN. This demonstrates the additional low-fidelity features does help improve the predictive capability of the network approximation. 

\begin{figure}[htbp]
    \centering
    \vbox{
    \includegraphics[scale=0.41]{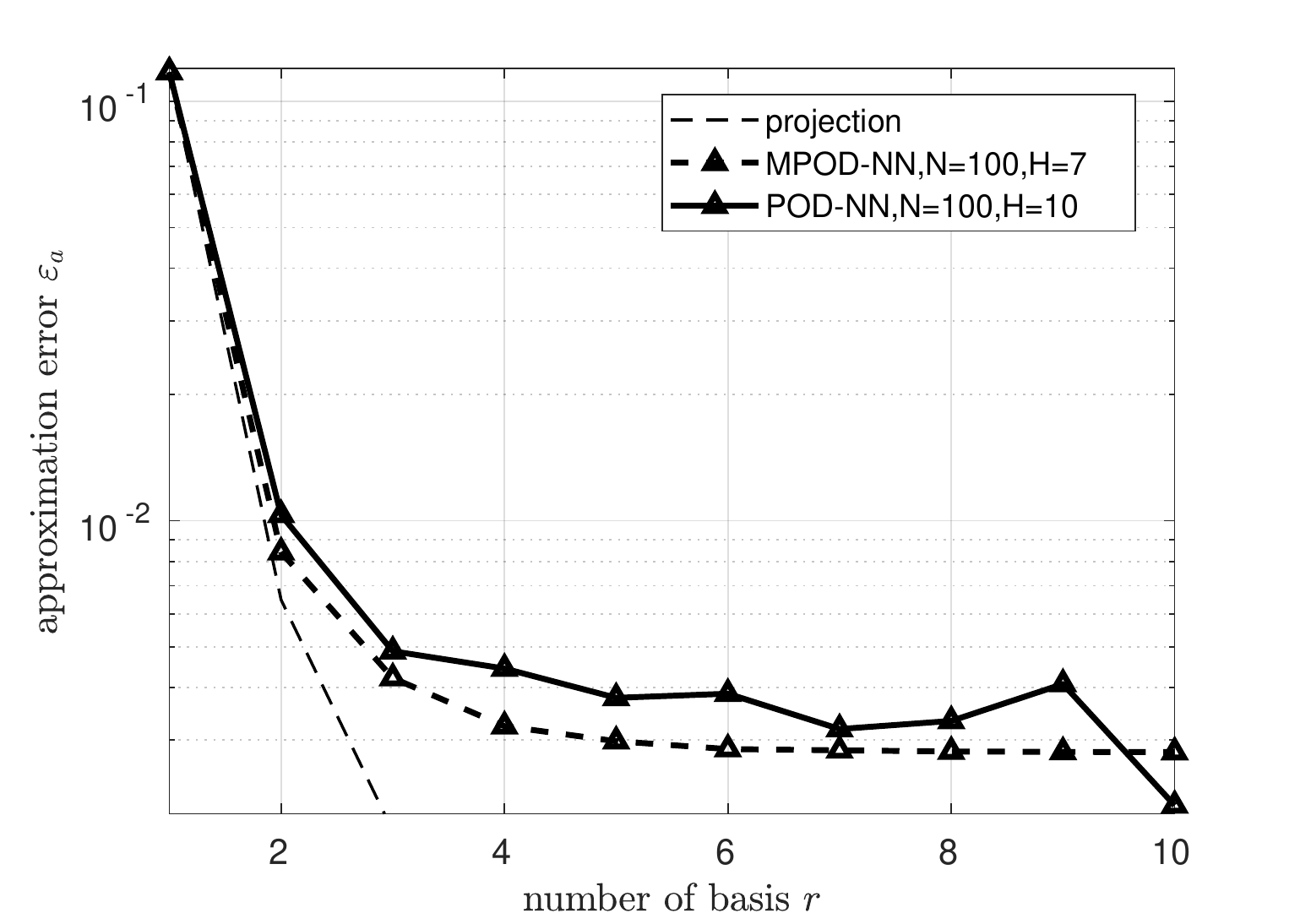}
    }
    \caption{Numerical convergence of approximation error $\varepsilon_a$ by original  POD-NN and modified POD-NN applied to problem \eqref{example9} with a training set of size $N=100$. The solid lines are the approximation errors of original POD-NN, and the dashed lines represent modified POD-NN.}
    \label{example9-comparison}
\end{figure}

 We also compare numerical convergence for the original POD-NN and the modified POD-NN  based on $100$ training points shown in Figure~\ref{example9-comparison}. The results suggest that both approaches reach a comparable accuracy for this example. In addition,  the hidden unit in each layer of   POD-NN  is $H=10$, while  the configuration of modified POD-NN is slightly more compact with $H=7$. These results justifies that the use of modified POD-NN as a baseline solution is reasonable as we mentioned in Section \ref{Modified POD-NN}.

\begin{figure}[htbp]
    \centering
    \vbox{
        \includegraphics[scale=0.42]{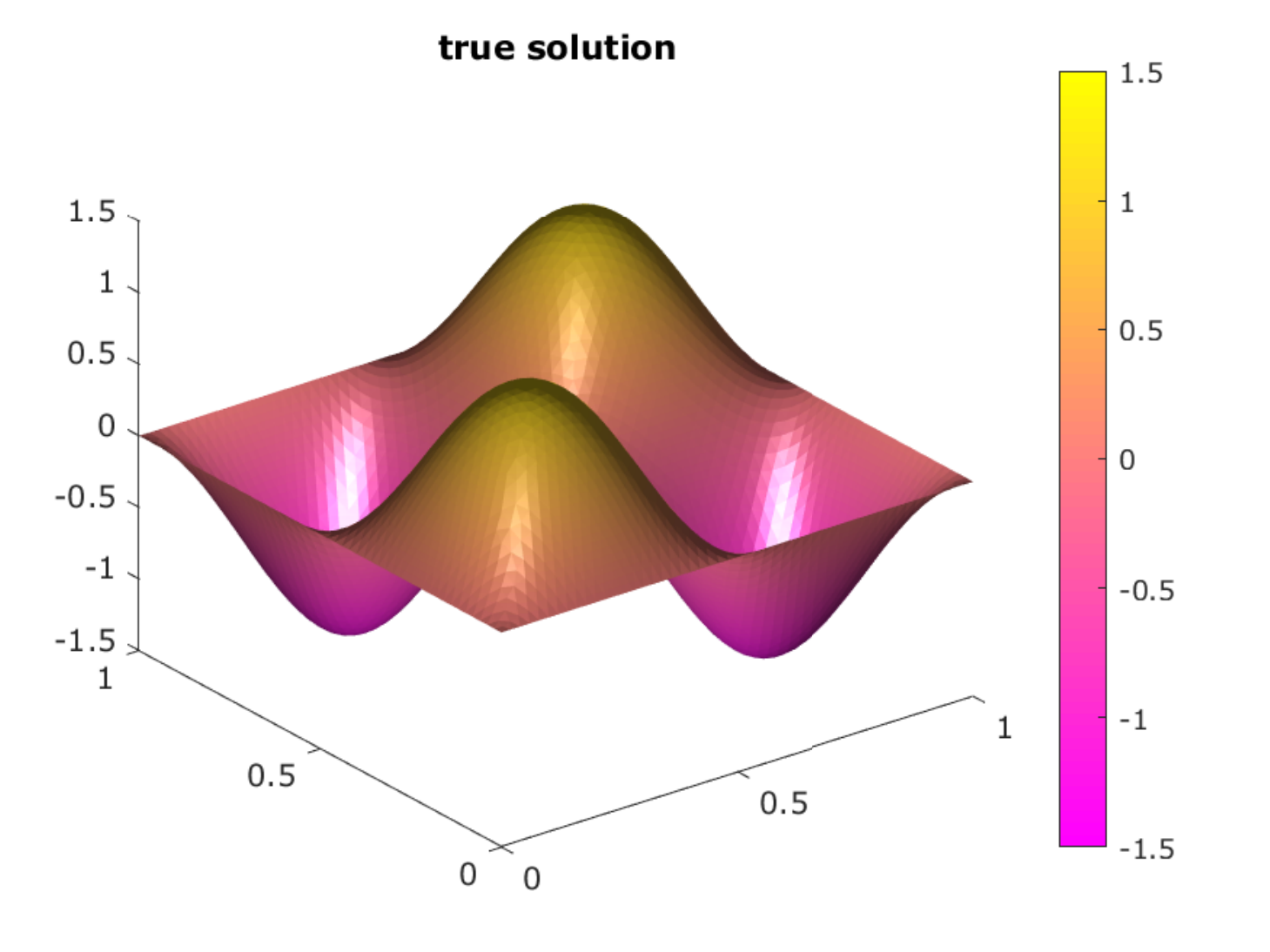}
        \includegraphics[scale=0.42]{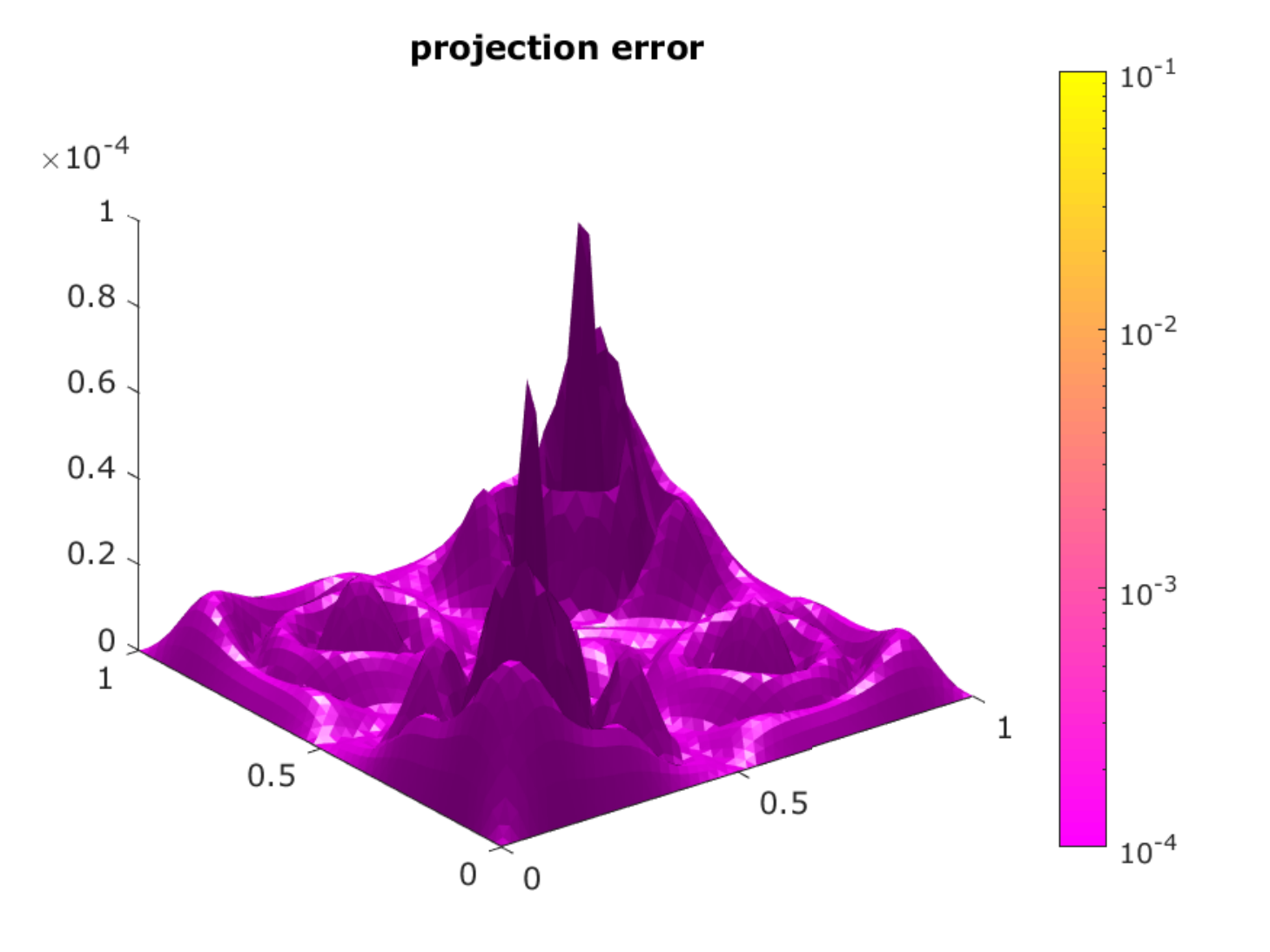}
    }
    \vbox{
        \includegraphics[scale=0.42]{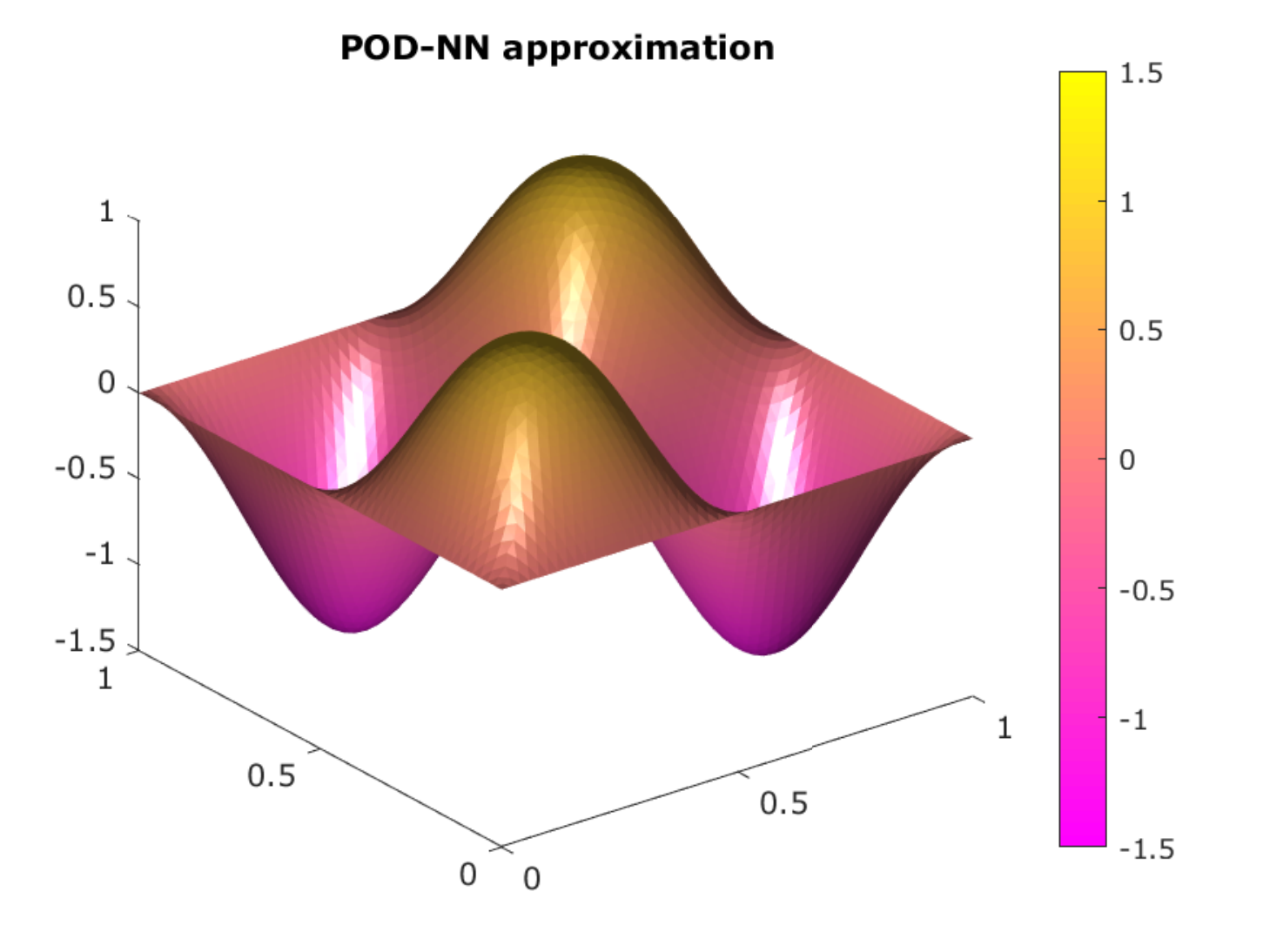}
        \includegraphics[scale=0.42]{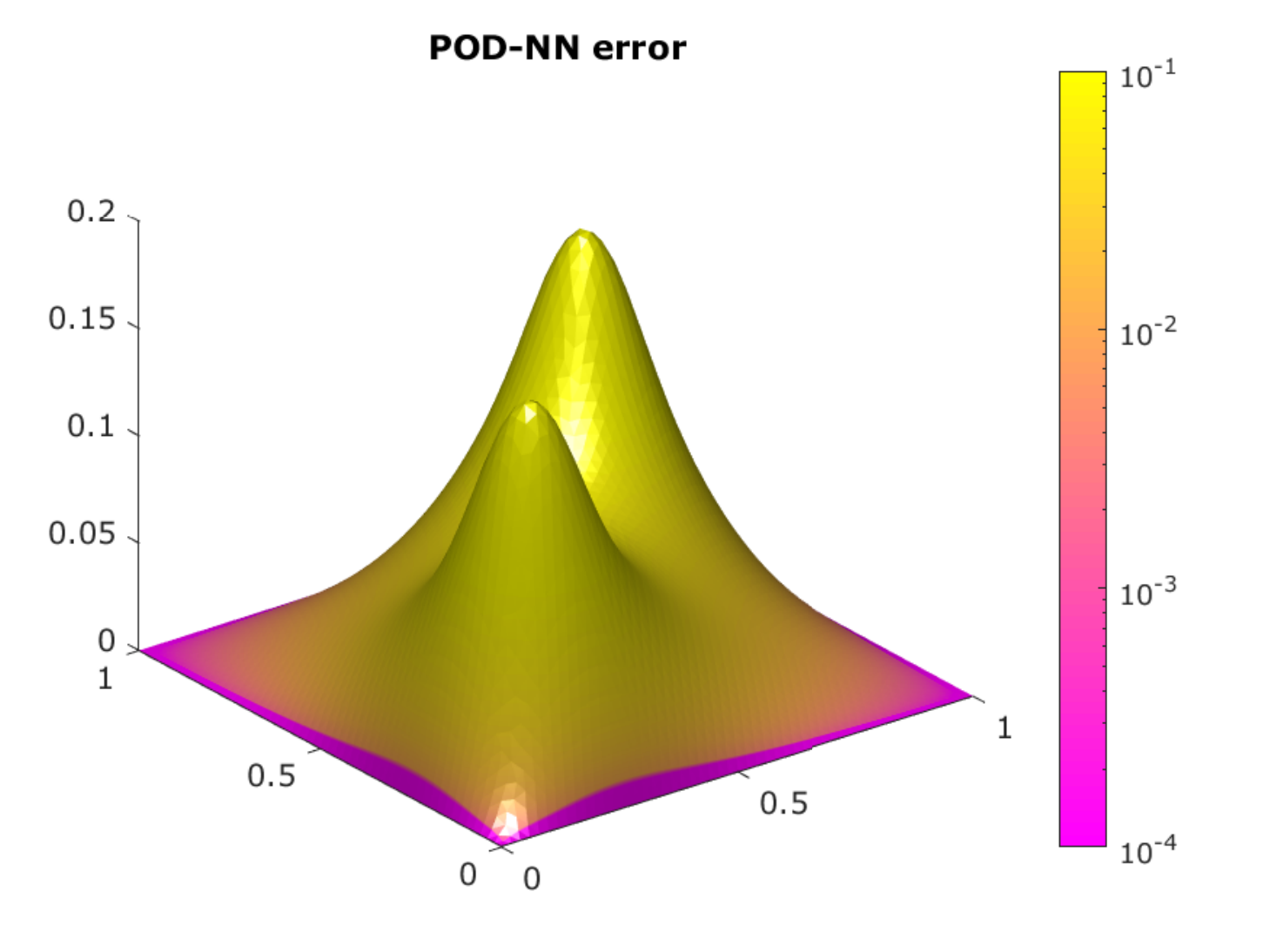}
    }
    \vbox{
        \includegraphics[scale=0.42]{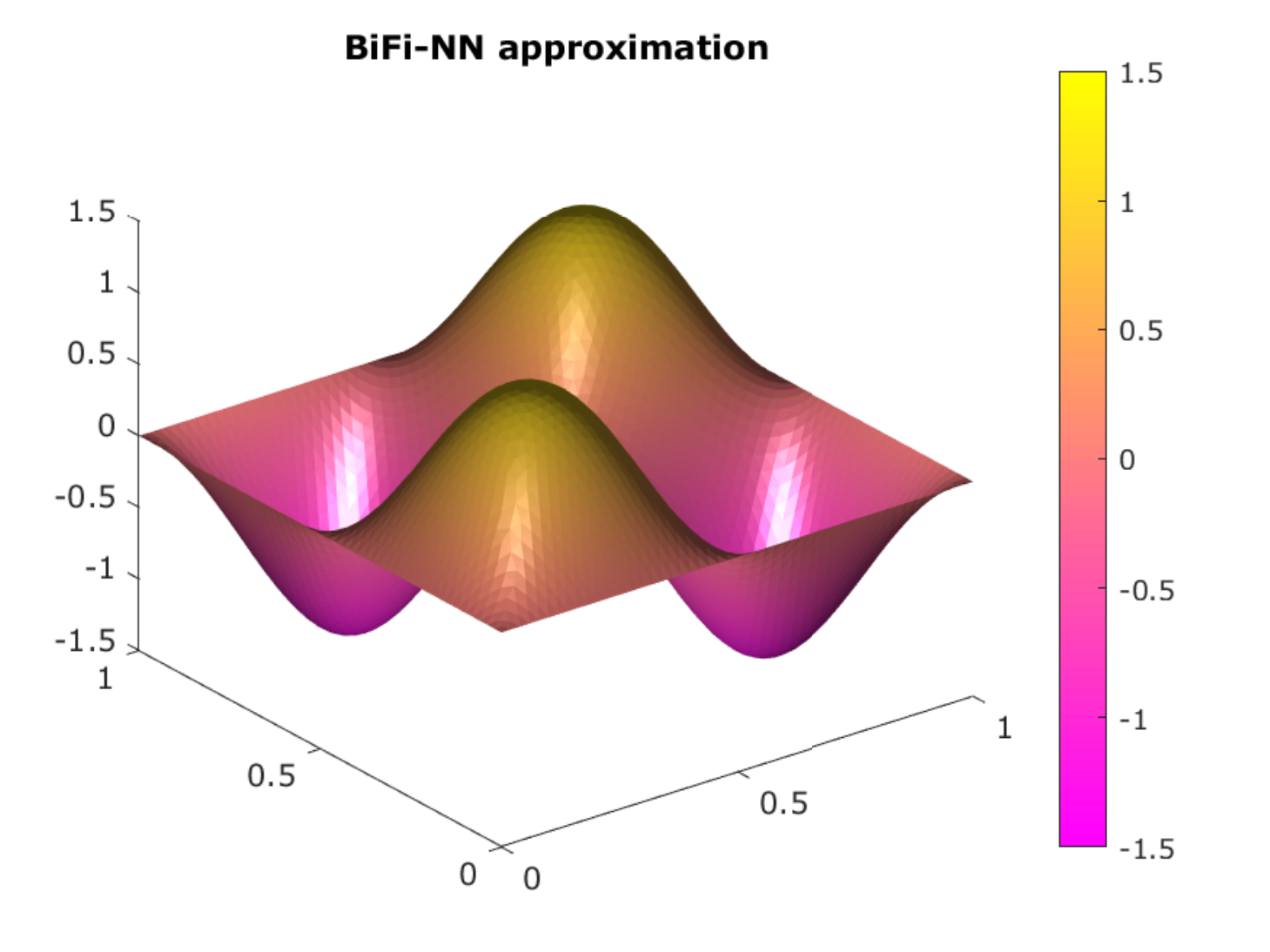}
        \includegraphics[scale=0.42]{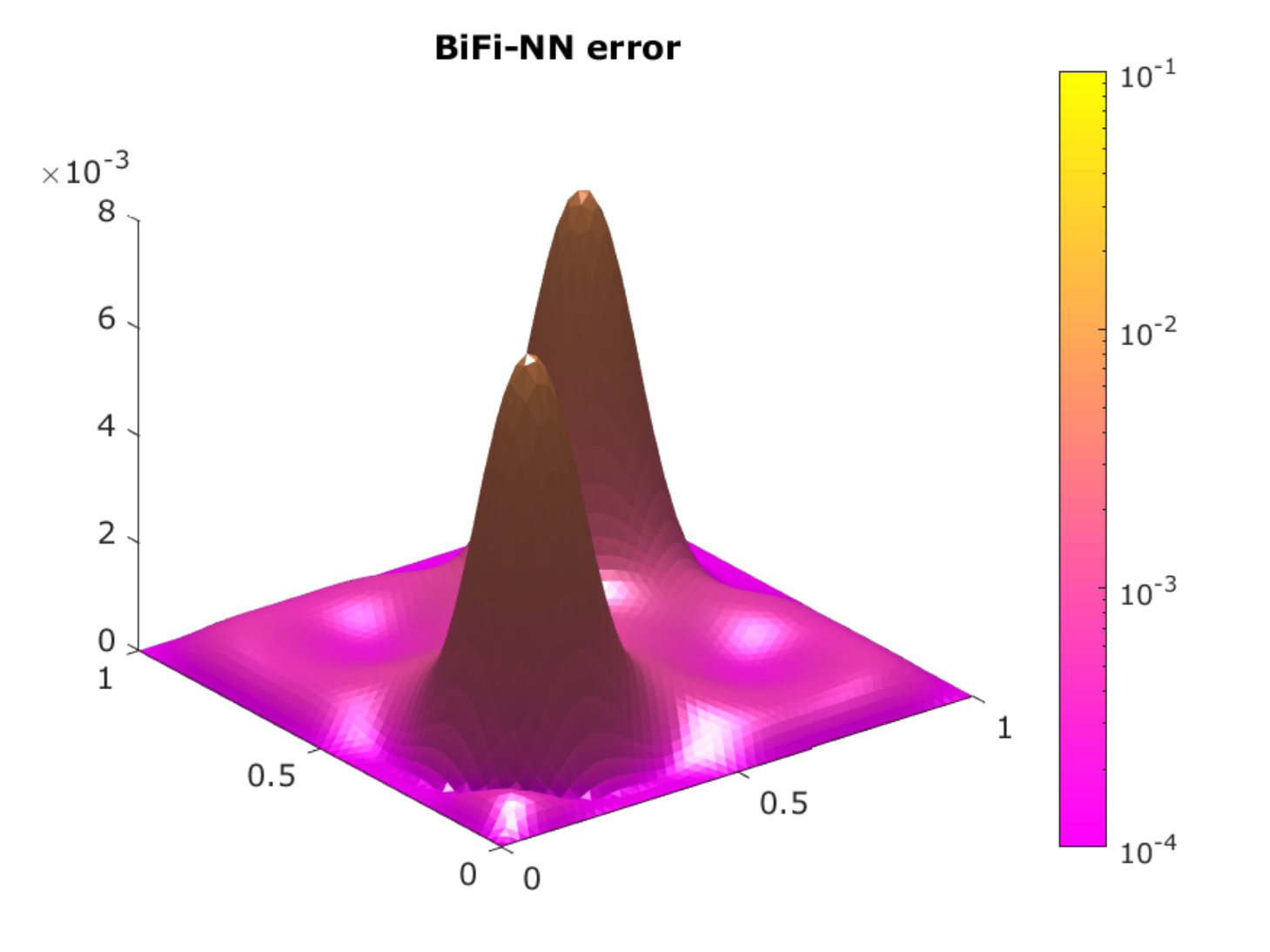}
    }
    \caption{The high-fidelity solution and the projection error (top), the approximation results and corresponding approximation errors by modified POD-NN (middle) and BiFi-NN (bottom) with $N=100$ training data and $r=10$ high-fidelity POD basis for $\mu = (4.928, 7.355).$}
    \label{example9-solution}
\end{figure}

To further demonstrate the accuracy of the BiFi-NN, we show the BiFi-NN and modified POD-NN approximation of  the high-fidelity solution for $\mu = (0.918, 0.010)$ in Figure \ref{example9-solution}. The first row is the  high-fidelity solution and the projection error (based on $r = 10$ POD basis), the second row shows the approximation and the corresponding approximation error by a pre-trained modified POD-NN model with $N=100$ training samples,  
while the last row is those based on BiFi-NN with the same training set. Both modified POD-NN and BiFi-NN show a good agreement with the high-fidelity solutions. 
However, BiFi-NN offered a  better accuracy, 
particularly around the peak of the solution. 

\subsection{2D Vorticity Equation}
\label{sec:2dns}
In the third example, we can consider the following  
2D vorticity equation  for an incompressible flow \cite{2dNS} with a random viscosity coefficient:
\begin{equation}
\begin{split}
    \label{example10}
    \partial_t w = \mu\Delta w-(u\cdot\nabla)w
\end{split}
\end{equation}
with the following initial condition:
\begin{equation}
    w|_{t=0} = \hat{w}+\epsilon(x, y),
\end{equation}
where
\begin{equation}
\begin{split}
    \hat{w}((x, y),0)&= \exp\left(-\frac{(x-\pi+\pi/5)^2+(y-\pi+\pi/5)^2}{0.3}\right) \\
    &\ -\exp\left(-\frac{(x-\pi-\pi/5)^2+(y-\pi+\pi/5)^2}{0.2}\right) \\
    &\ +\exp\left(-\frac{(x-\pi-\pi/5)^2+(y-\pi-\pi/5)^2}{0.4}\right), \\
\end{split}
\end{equation}
and $\epsilon(x, y)$ is a random noise uniformly distributed in [-1, 1] to the initial condition, which is fixed among all the training and test samples. The viscosity $\mu$ is varied in the range of $[2\times 10^{-3}, 5\times 10^{-3}]$, and the spatial variables $(x, y) \in [0, 2\pi]\times[0, 2\pi]$. 

Fourier spectral method is employed to solve this problem until the final time  $T = 50$ with a time step $\Delta t = 0.1$. The high-fidelity model is solved on a uniform grid of size $128\times 128$ in the spatial domain, while the low-fidelity model is solved on a coarser uniform mesh with a size of  $16\times 16$.  Training samples and 100  test samples are drawn independently by LHS. The reduced basis set is generated over an independent set of 100 sample points drawn by LHS.  

\begin{figure}[htbp]
\centering
\vbox{
\includegraphics[scale=0.42]{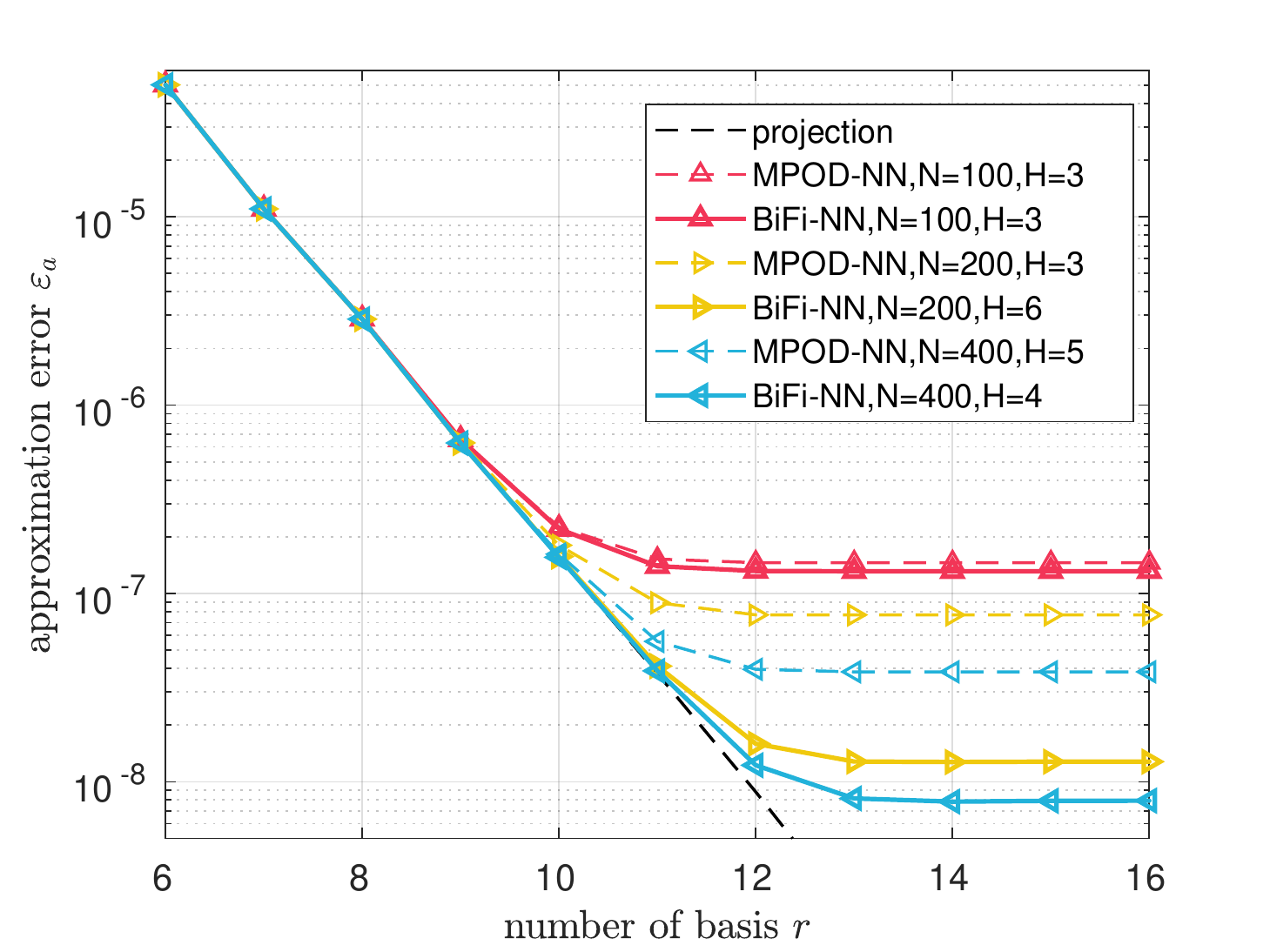}
\includegraphics[scale=0.42]{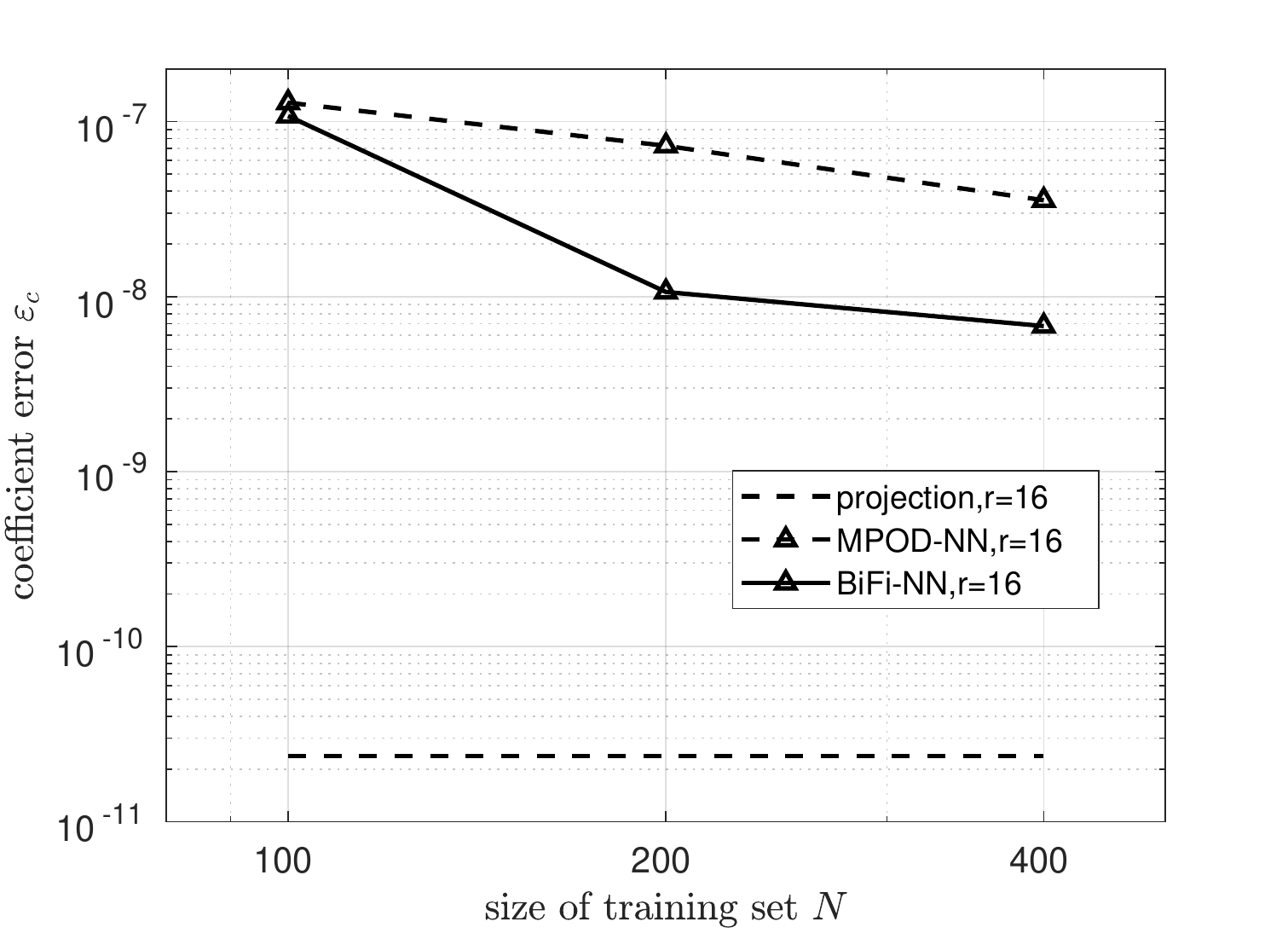}
}
\caption{Left: Numerical convergence of approximation error $\varepsilon_a$ by modified POD-NN and BiFi-NN applied to problem \eqref{example10} with training sets of different sizes ($N=100,200,400$). Right: Coefficient error with $r = 16$ by modified POD-NN and BiFi-NN with respect to the size of training set compared to the projection error. The solid lines are results of BiFi-NN, and the dashed lines represent modified POD-NN.}
\label{example10-errors}
\end{figure}

Figure \ref{example10-errors} (left) illustrates the approximation error convergence of both modified POD-NN and BiFi-NN methods with  the  number of the high-fidelity POD basis retained. Again, fast error decay for both methods is observed when the number of POD basis is small. When the reduced dimension $r$ is large enough, 
it is evident that BiFi-NN  delivers better results over modified POD-NN for a fixed train set. This signified the effectiveness of additional low-fidelity features we incorporate in BiFi-NN framework.  

Figure \ref{example10-errors} (right) presents  the coefficient errors with
respect to the size of the training set, when the dimension of the  reduced approximation space is fixed at $r=16$. The coefficient errors are roughly 10 times larger
than the projection error, indicating that the dominant error contribution is due to the coefficient error committed by the neural network approximation. It is clear that by utilizing the information from the low-fidelity model, BiFi-NN is able to improve the accuracy of approximation of the high-fidelity reduced coefficients and offer more accurate reduced solutions.

\begin{figure}[htbp]
    \centering
    \vbox{
    \includegraphics[scale=0.6]{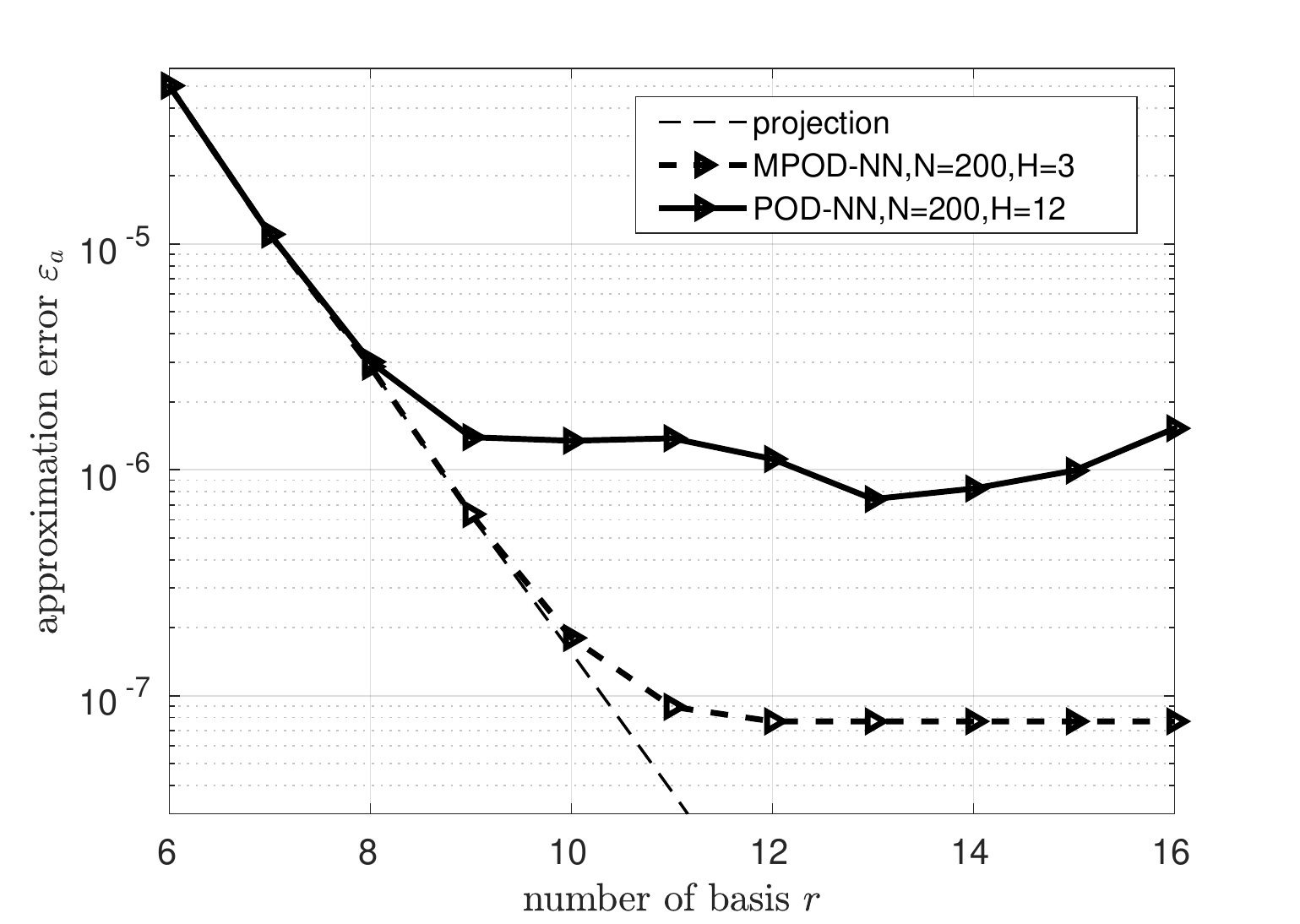}
    }
    \caption{Numerical convergence of approximation error $\varepsilon_a$ by original POD-NN and modified POD-NN applied to problem \eqref{example10} with a training set of  size $N=200$.  The solid lines are the approximation errors of original POD-NN, and the dashed lines represent modified POD-NN.}
    \label{example10-comparison}
\end{figure}

 We also compare numerical convergence for the original POD-NN and the modified POD-NN based on $200$ training points  shown in Figure~\ref{example10-comparison}. In this example, the modified POD-NN produced better results 
as the reduced dimension $r$ is large enough for this example. 
This might be because the network configuration of the modified POD-NN ($H=5$) is more compact with that of the original POD-NN ($H=13$). Therefore, it is easier to train with the same training set.


\subsection{2D flow around cylinder} In the last example, we consider a 2D  channel flow around a cylinder \cite{larson2013finite} with a random inflow condition, which is modeled by the following Naiver-Stokes equations: 
\begin{equation}
\label{example11}
    \begin{aligned}
    \dot{u} + (u\cdot \nabla)u+\nabla p-\nu\Delta u &= f, \ &\text{in}\ \Omega\times (0,T], \\
    \nabla\cdot u &= 0, \ &\text{in}\ \Omega\times (0,T], \\
    u &= g_{D}, \ &\text{in}\ \Gamma_{D}\times (0,T], \\
    \nu n\cdot \nabla u-pn &= 0, \ &\text{in}\ \Gamma_{N}\times (0,T], \\
    u &= u_0, \ &\text{in}\ \Omega,\quad \text{for} \quad {t=0},
    \end{aligned}
\end{equation}
where $u$ and $p$ are the sought velocity and pressure, and $f=1$ is a given body force \cite{larson2013finite}.  The fluid has viscosity $\nu = 10^{-3}$ and unit density. The problem is defined on a channel $\Omega \in [0,2.2]\times [0.41]$, with a cylinder of diameter $0.1$ centered at $(0.2, 0.2)$. The boundary $\partial \Omega$ is divided into two parts $\Gamma_D$ and $\Gamma_N$, where $\Gamma_D$ denotes either the rigid walls of the channel with $g_D = 0$, or the inflow region with $g_D$ the inflow velocity profile, and $\Gamma_N$ denotes the outlet.
On both upper and lower wall and on the cylinder, a non-slip boundary condition is prescribed. On the right wall, zero initial conditions are assumed. A random inflow profile with  $U_{\max}=0.3$ is given on the left wall:
\begin{equation}
    u_{in} = \frac{4U_{\max}y(0.41-y)}{0.41^2}a(y, z),
\end{equation}
with
\begin{equation}
    a(y, z) = 1+\sigma\sum_{k=1}^{d}\frac{1}{k\pi}\cos(2k\pi y)z_k,
\end{equation}
where $z = (z_1,\hdots,z_d)\in [-1, 1]^d$ subjects to a random uniform distribution. 
Here we fix $d = 5$ and $\sigma = 1$ (where the positivity of $a(y,z)$ is guaranteed).

\begin{figure}[htbp]
    \centering
    \includegraphics[scale=0.8]{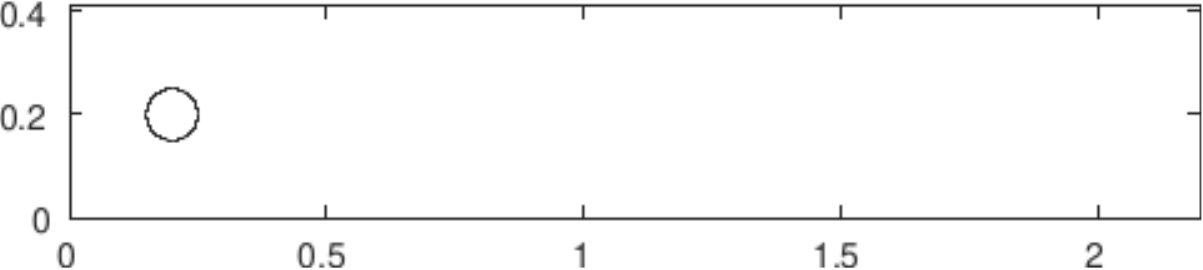}
    \caption{The geometry for the 2D flow past cylinder problem \eqref{example11}.}
    \label{example11-mesh}
\end{figure}

We solve the above equation with $P_1$ finite elements. Similar to the previous example, we consider the following two models: low-fidelity solutions use 119 elements and high-fidelity solutions use 2522 elements. The geometry and mesh for the low-fidelity model is illustrated in Figure \ref{example11-mesh}. The time step is set to be $\Delta t = 0.01$ and the final time is $T = 1.0$. Our output of interest is the magnitude  of the flow field. In this problem, 100 Monte Carlo samples in the parameter space are employed as the test set and an independent set of 100 Monte Carlo samples is utilized to compute the reduced basis set. 

\begin{figure}[htbp]
\centering
\vbox{
\includegraphics[scale=0.42]{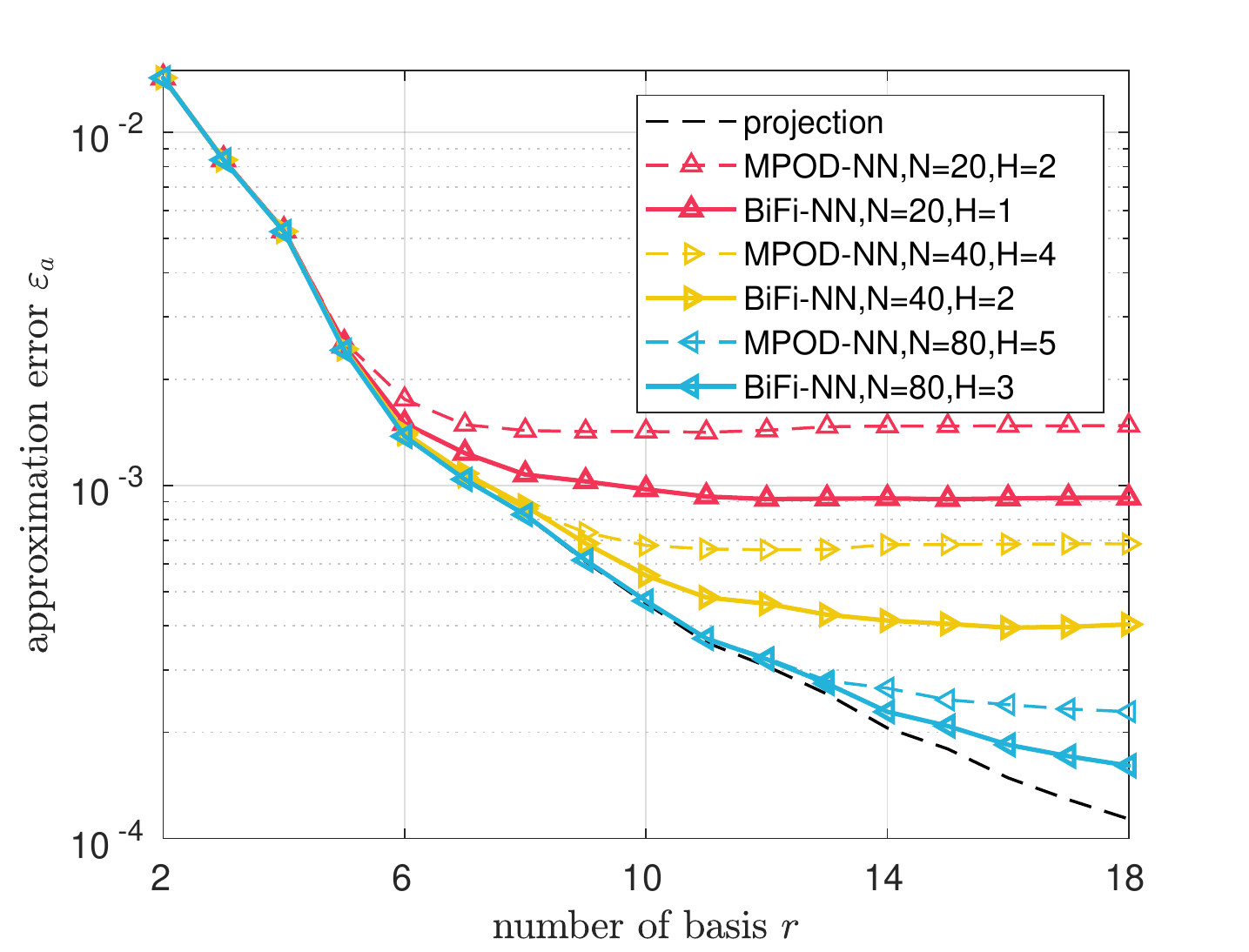}
\includegraphics[scale=0.42]{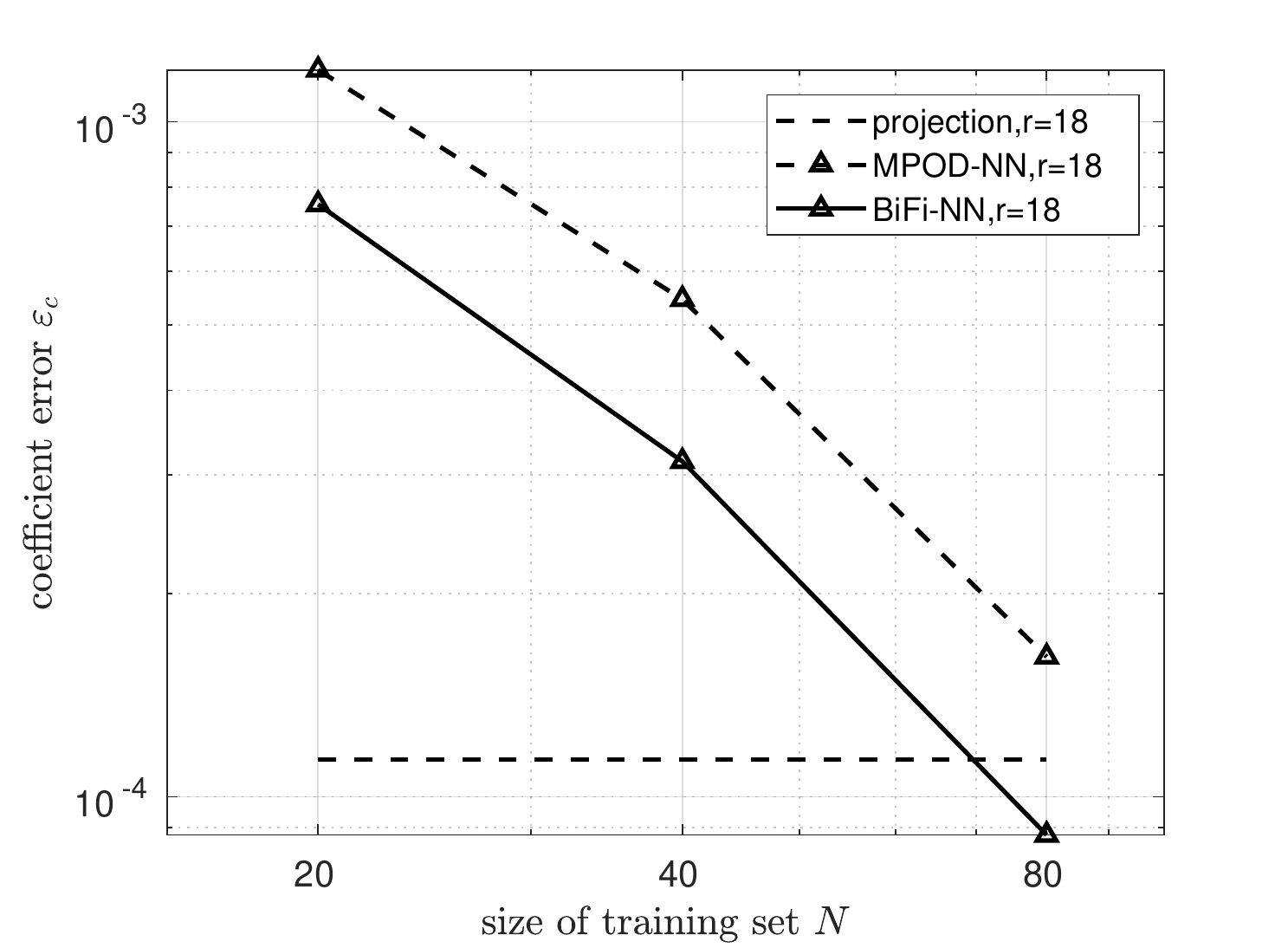}
}
\caption{Left: Convergence analysis of approximation error $\varepsilon_a$ by modified POD-NN and BiFi-NN for problem \eqref{example11} with training sets of different sizes ($N=20,40,80$). Right: Coefficient error with $r = 18$ by Modified POD-NN and BiFi-NN with respect to the size of training set compared to the projection error. The solid lines are results of BiFi-NN, and the dashed lines are of modified POD-NN.}
\label{example11-errors}
\end{figure}

The approximation errors of both modified POD-NN and BiFi-NN based on three different training sets ($N=20,40,80$) are plotted in Figure \ref{example11-errors} (left). We observed a fast decay for both methods and they both stagnate roughly around $\mathcal{O}$(10) high-fidelity POD basis, indicating the coefficient error begins to dominate over the projection error. The convergence behaviors of modified POD-NN and BiFi-NN are similar. However, BiFi-NN can continue to decrease and saturates at a lower error level for a fixed training set. When more training data is available, the error saturation  level  can be further reduced shown in Figure \ref{example11-errors} (left).

The coefficient errors of both methods with respect to the number of training points, for a fixed reduced dimension $r=18$ is further analyzed in  Figure \ref{example11-errors} (right). In this case, the coefficient error is dominant over the project error, particularly for the small training set. When the size of training data increases, the coefficient errors of both methods can be further improved. It is evident that compared with modified POD-NN,  the improvement on the approximation of the high-fidelity coefficients is quite noticeable by  incorporating the low-fidelity feature on BiFi-NN.


\section{Summary}
\label{sec:summary}

In this paper, we proposed a new nonintrusive reduced-order modeling method (referred to as BiFi-NN).  With both low-fidelity data and high-fidelity data, the method generates the reduced basis set from the collection of high-fidelity snapshots by POD, and employs two-hidden-layer perceptron to approximate the high-fidelity POD  coefficients with both low-fidelity POD coefficients and the physical model parameters  as input features. With an affordable computational cost, we demonstrated the improved predictive performance of the proposed method and the effectiveness of  the additional features extracted from low-fidelity models via several benchmark examples, particularly for nonlinear problems. 
Future work includes  evaluating the framework on more complex problems and extending this idea to general multi-fidelity case, i.e., the number of fidelities is larger than three.



\section{Acknowledgement}

We thanks for the Simons Foundation (504054) for their funding support.


\bibliographystyle{abbrv}
\bibliography{collocation,random,NN}


\end{document}